\documentclass[11pt]{article}

\usepackage{algorithm}
\usepackage{algpseudocode}
\usepackage{booktabs}
\usepackage{tabularx}
\usepackage{placeins}   
\usepackage{float}      
\usepackage{url}
\usepackage{authblk}
\usepackage{graphicx}

\usepackage{authblk}   
\usepackage[hidelinks]{hyperref}

\date{}

\usepackage{amssymb}
\usepackage{amsmath}

\title{\textbf{X-Blocks: Linguistic Building Blocks of Natural Language Explanations for Automated Vehicles}}

\author[1,2]{Ashkan Y. Zadeh}
\author[1,2]{Xiaomeng Li}
\author[1,2]{Andry Rakotonirainy}
\author[1,2]{Ronald Schroeter}
\author[1,2]{Sebastien Glaser}
\author[1,2]{Zishuo Zhu}

\affil[1]{Queensland University of Technology (QUT), Brisbane, Australia}
\affil[2]{ARC Training Centre for Automated Vehicles in Rural and Remote Regions (AVR3), Brisbane, Australia}

\begin{document}

\maketitle

\begin{abstract}
Natural language explanations play a critical role in establishing trust and acceptance of automated vehicles (AVs), yet existing approaches lack systematic frameworks for analysing how humans linguistically construct driving rationales across diverse scenarios. This paper introduces X-Blocks (eXplanation Blocks), a hierarchical analytical framework that identifies the linguistic building blocks of natural language explanations for AVs at three levels: context, syntax, and lexicon. At the context level, we propose RACE (Reasoning-Aligned Classification of Explanations), a multi-LLM ensemble framework that combines Chain of Thought reasoning with Self Consistency mechanisms to robustly classify explanations into 32 scenario-aware categories. Applied to human-authored explanations from the Berkeley Deep Drive-X dataset, RACE achieves 91.45\% accuracy and Cohen's kappa of 0.91 against human annotations, demonstrating almost perfect inter-rater agreement. At the lexical level, we employ log-odds analysis with informative Dirichlet priors to extract context-specific vocabulary patterns that distinguish driving scenarios. At the syntactic level, dependency parsing and template extraction reveal that explanations draw from a limited repertoire of reusable grammar families, with systematic variations in predicate types and causal constructions across contexts. The X-Blocks framework is dataset-agnostic and task-independent, offering broad applicability to other AV datasets and safety-critical domains. Our findings provide evidence-based linguistic design principles for generating scenario-aware explanations that pave the road to enhance user trust and cognitive accessibility in automated driving systems.
\end{abstract}

\section*{Keywords}
Automated Vehicles; Natural Language Processing; Explainable Automated Vehicles; Human--Vehicle Interaction; Human-Centred Artificial Intelligence



\section{Introduction}

\subsection{Background and Motivation}
The advent of automated vehicles (AVs) has led to a transformative era in transportation and is promising significant advancements in road safety, traffic efficiency and sustainability \cite{mehraban2024fuzzy,zadeh2024integrated}. However, it is critical that these vehicles not only perform with technical robustness, but also provide clear, human-centred explanations of their actions to users if they are to be widely adopted and truly trustworthy \cite{kuznietsov2024explainable,tekkesinoglu2024advancing}. Understanding the rationale behind an AV’s decision-making is critical for enhancing user trust, fostering acceptance, facilitating smooth control transition, and ensuring timely human responses in safety-critical driving scenarios. In turn, this can help mitigate risks associated with misuse, unsafe or unexpected automated behaviours \cite{yousefizadeh2025psylingxav,zhu2025humancentric}. 

Artificial intelligence (AI) lies at the core of an AV’s decision-making, primarily through sophisticated techniques such as Machine Learning (ML), Deep Learning (DL) \cite{mehraban2025saliency}, and transformer-based models \cite{chen2024endtoend}. However, the complexity of these AI systems creates a significant challenge: they often operate as "black boxes", unable to provide clear explanations for their decisions, which can leave users feeling uncertain or sceptical \cite{voneschenbach2021transparency}. This lack of transparency, combined with the non-linear nature of AI decisions and the algorithm's disregard for human factors, is a significant concern within the Explainable Artificial Intelligence (XAI) community\cite{nascimento2020systematic}. Furthermore, current data-driven methods often face challenges with data dependency, poor generalisation, and difficulties in handling rare scenarios due to the high cost of data collection and annotation \cite{ding2023survey}. Recent advancements of Large Language Models (LLMs) have opened a new route for addressing these critical challenges in automated driving and transportation research \cite{chen2024endtoend,yang2023llm4drive}. LLMs, exemplified by models like GPT-4, possess remarkable capabilities in semantic comprehension, logical reasoning, knowledge generalisation, and generate human-like text \cite{yang2023llm4drive}. This ability to generate human-like natural language explanations is particularly crucial for advancing XAI and fostering trustworthy AVs \cite{yousefizadeh2025psylingxav}.

\subsection{Problem Statement and Research Gap}
\label{subsec:Problem}

\begin{figure}[t]
    \centering
    \includegraphics[width=\textwidth]{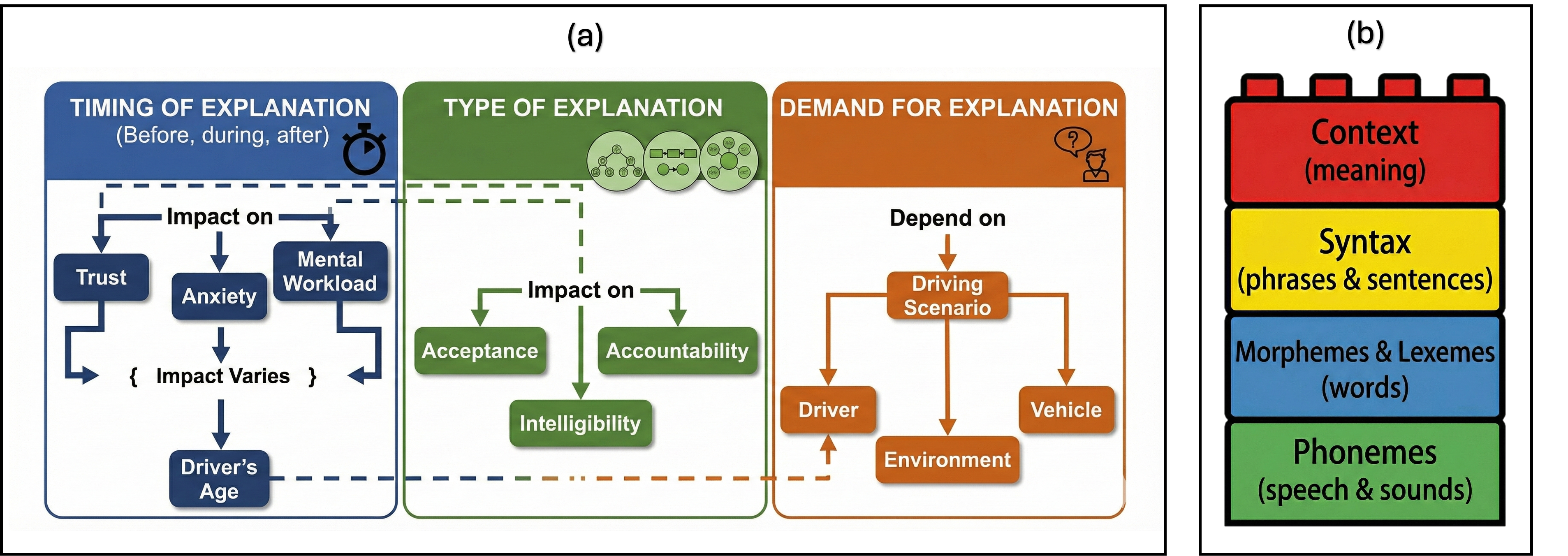}
    \caption{(a) Key factors affecting the effectiveness of explanations in AVs: Insights from user-centered studies~\cite{yousefizadeh2025psylingxav} and (b) Building blocks of language~\cite{vajjala2020practical}.}
    \label{FIG:lang_blocks}
\end{figure}

Our research in Psycholinguistic Language Explanation for AV (PsyLingXAV)  highlighted that the type and demand for explanations in AVs are not uniform, but highly dependant on contextual factors, especially the driving scenario in which a behaviour occurs ~\cite{yousefizadeh2025psylingxav}. As illustrated in PsyLingXAV and shown in Figure~\ref{FIG:lang_blocks}\,(a), explanation needs vary based on the interaction between drivers, vehicles, and their surrounding environments, with scenario context playing a central role in shaping user expectations and information needs. This underscores the critical importance of scenario-aware explainability for achieving intelligible, trustworthy, and user-aligned AV behaviour~\cite{yousefizadeh2025psylingxav}. Human drivers routinely justify their actions with reference to situational cues, such as traffic flow, road geometry, pedestrian behaviour, or weather conditions. Recent evidence from Zhu et al.~\cite{zhu2025humancentric} further reinforces that human-centric explanations must be tailored to specific driving contexts, since a one-size-fits-all approach often leads to misunderstanding and degraded trust. They revealed that most prior work applied the same structure and timing of explanations across different scenarios, neglecting contextual diversity and failing to capture how users interpret AV behaviour under real-world complexity.

To systematically address the structure of such context-dependant explanations, we draw on our linguistic framework illustrated in Figure~\ref{FIG:lang_blocks}\,(b). The building blocks of language provide a principled framework for designing context-sensitive AV explanations. The \textbf{Context} layer corresponds to the driving scenario, providing the semantic grounding that determines \textit{what} needs to be explained. The \textbf{Syntax} and \textbf{Morphemes/Lexemes} layers relate to the type of explanation, governing \textit{how} explanations are structurally organised and \textit{which} vocabulary is selected to achieve intelligibility for different users. While the \textbf{Phonemes} layer and explanation timing fall outside this research's scope, Figure~\ref{FIG:lang_blocks}\,(a) shows that timing is intrinsically linked to both type and demand, underscoring their interconnected nature. This framework ensures that explanations are not only factually correct but also contextually appropriate and cognitively accessible.

Nonetheless, there is a profound research gap in the area of explainable AVs. Current datasets of driving-action explanations are not designed to systematically relate linguistic choices to contextual requirements. Explanations are typically considered as holistic outputs, or broken down into broad categories like “action” and “justification”, without specific attention to how explanations should vary in structure, content, and level of detail depending on the driving context~\cite{dong2023why}. For instance, the Berkeley Deep Drive-X (BDD-X) dataset records human-authored explanations at scale~\cite{kim2018textual}, but lacks scenario-level labels to contextualise explanations to specific needs in a given driving situation. Thus, there is a lack of contextually informed linguistic building blocks to analyse how humans naturally communicate situation-specific driving explanations and to develop AV explanations that are understandable, trustworthy, and aligned with user needs for different driving scenarios.

\subsection{Research Questions}
Given the lack of systematic mapping between natural language driving explanations and their underlying linguistic structures, this research investigates the fundamental language building blocks that constitute human explanations of driving actions. We propose X-Blocks, an analytical framework that deconstructs textual driving explanations into hierarchical linguistic levels: scenario-level context categorisations, lexeme-level morphological patterns, and syntax-level structural compositions. This multi-level approach enables systematic analysis of how humans linguistically encode driving explanations, establishing a foundation for scenario-aware AV explanation generation. Accordingly, this research addresses three questions:
\textit{RQ1.} How can natural language driving explanations be systematically categorised into scenario context blocks?
\textit{RQ2.} What are the morphological and lexical patterns of explanations across different driving contexts?
\textit{RQ3.} What are the syntactic structures of explanations across different driving contexts?


\subsection{Research Contributions}

This study advances explainability in AVs research by introducing a generalisable, linguistically grounded framework for analysing and designing natural language explanations of driving behaviour. While human explanations could exhibit considerable variability, the proposed approach identifies dominant recurring patterns at the levels of context, syntax, and lexicon that can inform explanation system design. These findings offer practical value for AV users, industry practitioners, and researchers in AVs and XAI. This work makes four contributions: \textit{C1.} We introduce a comprehensive taxonomy of driving contexts that enables systematic interpretation and evaluation of natural language explanations with respect to the dominant situational factors underlying driving behaviour, supporting clearer and more relevant AV communication across diverse scenarios. \textit{C2.} We propose X-Blocks, a hierarchical framework that decomposes explanations into scenario context, syntactic structure, and lexical choice. The framework is dataset-agnostic and task-independent, making it applicable to other AV datasets, explanation-generation systems, and broader safety-critical domains requiring structured analysis of explanatory language. \textit{C3.} We present the RACE framework, the first level of X-Blocks, which combines Chain of Thought(CoT) reasoning with self consistency(SC) across multiple large language models to robustly classify explanations into scenario contexts. This approach enables scalable and transparent annotation of explanation corpora, reducing dependence on manual labelling and supporting industrial deployment. \textit{C4.} Through systematic lexical and syntactic analysis, we characterise how explanatory language varies across driving contexts, providing actionable linguistic design insights for generating context-sensitive explanations that enhance user trust and cognitive accessibility.

\section{Related Works}

\subsection{Driving Scenario and Context Classification}
Driving contexts are inherently complex and multi-factorial.  Human driver or AV action in the real world often stems from more than one contributing cause. When explanations describe multiple factors, identifying the primary causal factor, which is the one that would be the main cause of the action, is difficult but essential \cite{li2024exploring}. This need becomes especially urgent in real-world AVs, where explanations must often be delivered under time constraints. According to previous research \cite{yousefizadeh2025psylingxav,zhu2025humancentric}, explanations given during or shortly before actions are more cognitively demanding and time sensitive. Therefore, focusing on the most important cause rather than listing all contributing factors is essential for minimising mental workload, reducing anxiety, and maintaining user trust \cite{yousefizadeh2025psylingxav,zhu2025humancentric}. Despite efforts to improve explainability, current models often struggle to isolate the dominant causal influence from confounding context \cite{giamattei2024causality}. Identification of such causal factors could lead to more explicit and structured scenario categorisation, improve the fidelity of behaviour models, and enhance downstream tasks such as benchmarking, safety validation, and behavioural analysis \cite{zhao2024risk}. Driving context classification has received substantial attention in recent years, with increasing focus on capturing inter-driver variability, safety assessment, and energy efficiency. Most existing approaches rely on sensor-derived data and vehicle kinematics, applying various unsupervised and supervised machine learning techniques to segment and cluster driver behaviours into interpretable categories. Several studies have explored clustering of driving styles using sensor and temporal data. Feng et al. \cite{feng2023driving} introduced a deep temporal clustering method that segments driving styles within trips, enhancing explainability via SHAP-based feature attribution. While early clustering approaches primarily relied on GPS or CAN bus signals, newer work has shifted toward latent topic models and behaviour annotation. 

More recent contributions integrate linguistic or explanation-centred models. Wang et al. \cite{wang2023goaldriven} proposed a goal-driven explainable clustering model that generates natural language descriptions of latent clusters, offering a pathway to bridging raw sensor data with semantic interpretation. However, these methods are concerned with the task of producing language from data, without exploring the other side of the coin, which is systematically categorising existing natural language explanations into structured driving contexts. This is a crucial step in comprehending how humans linguistically represent driving rationale. Furthermore, the study by Chen et al.\cite{chen2023feature} on clustering driving style and skill with naturalistic data and behavioural questionnaires illustrates that the inclusion of human self-reporting data can complement sensor data features for improved clustering results. However, this method does not, either, explore the analysis of natural language explanations in the context of driving scenarios. While recent developments have shown the capability of performing complex clustering and classification tasks using sensor data, and some have explored the latent language mapping concept, none of the reviewed studies have explored the classification of natural language driving explanations into structured context-aware categories. This is why there is a need for a systematic approach to map natural language driving explanations to scenario-based context categories, as described in Section ~\ref{subsec:Problem}.

\subsection{Language Processing and Generating in AVs}
\subsubsection{Large Language Models in Automated Vehicle Research}
Large Language Models (LLMs) are transformer models that are trained on large text corpora, allowing them to perform text understanding, generation, and reasoning tasks \cite{vaswani2017attention, zhao2023survey, minaee2024large}. In the context of research on AVs, LLMs and Vision-Language Models (VLMs) have been used mainly in two areas: (1) explanation generation from driving scenes, and (2) incorporating language understanding into end-to-end driving systems.
In the first category, models like ADAPT \cite{jin2023adapt} and DriveGPT4 \cite{xu2023drivegpt4} analyse video inputs to generate text descriptions of driving actions and explanations. This category views language as an output, where machine-authored explanations are created to improve explanation. In the second category, models like LLM-Driver \cite{chen2023driving} and LanguageMPC \cite{sha2023languagempc} apply LLMs for understanding and motion planning, which show excellent capabilities in causal reasoning \cite{mao2023gptdriver,wen2023ontheroad}.
However, these methods have one thing in common: they all highlight machine-produced language for explanation or control. Relatively less effort has been put into the analysis of human-produced explanations, which is the natural language used by drivers or annotators to describe their own decisions while driving. Analysis of human-produced explanations is important for human-centric AV development, as it helps in understanding how humans think about and explain driving contexts \cite{albrecht2022despite, liao2023ai}. This research fills this gap by using LLMs not for explanation generation but for the classification of human-authored driving explanations into pre-defined contextual categories based on a domain-specific taxonomy.

\subsubsection{Chain of Thought and Self Consistency Reasoning}
Chain of Thought prompting is a method of prompting LLMs to reason step by step towards a final answer, instead of directly producing an output \cite{wei2022chain, alammar2024hands}. This is a reflection of human deductive reasoning and has been successful in tasks that involve structured reasoning. CoT has been successful in classification and annotation tasks in various domains such as producing labelling rationales for medical and legal annotation \cite{lee2024applying}, simulating human-like decision-making for disease classification \cite{pan2025chatleafdisease}, and assessing free-text answers \cite{wang2022self}.
The drawback of the conventional CoT is that it may have inconsistent reasoning chains when using greedy decoding. self consistency is proposed to tackle this issue by sampling multiple reasoning chains and choosing the most frequent or consistent response based on majority voting \cite{wang2022self, alammar2024hands}. Variants of SC, such as Soft self consistency, enable weighted agreement among the responses, which is very helpful in the case of ambiguous classification problems in which human annotators may also have inconsistent opinions \cite{wang2024soft}.
These characteristics make CoT-SC an appropriate approach for classifying human-authored driving explanations. These driving explanations are necessarily variable, which means a given explanation can refer to more than one context, and a given text can be interpreted in more than one way depending on the reasoning method. CoT allows for systematic reasoning about explanations in driving context. SC adds robustness by combining the results of multiple instances of a model to account for variability.

\subsubsection{Prompt Engineering with CoT for Automated Vehicles}
There has been some recent research that has started applying CoT reasoning to AV systems, mostly for decision-making and planning. PKRD-CoT proposes a framework based on Perception, Knowledge, Reasoning, and Decision-making, allowing for zero-shot CoT prompting in multimodal driving scenarios \cite{luo2024pkrdcot}. Agent-Driver models LLMs as cognitive agents with memory, applying CoT reasoning to traceable motion plans \cite{mao2023language}. The Receive–Reason–React framework links driving decisions to natural language user inputs via CoT prompting \cite{cui2024receive}, and DriveRP integrates CoT with retrieval-augmented generation for improved situational awareness \cite{huang2024driverp}. In addition to planning, CoT has been used for classifying crash severity based on narrative descriptions, showing the effectiveness of CoT for language-based reasoning in transportation studies \cite{zhen2024leveraging}.
These experiments show that CoT improves the interpretability and enables common-sense reasoning in AV scenarios. Nevertheless, the current state-of-the-art applications are mostly related to machine reasoning for planning/control tasks, without paying much attention to the classification of natural language explanations provided by humans.
This research fills this gap by using CoT-SC reasoning to classify human-authored driving explanations from the BDD-X dataset into context-specific classes. We hypothesise that this is an effective method for two reasons. Firstly, driving explanations require implicit multi-step reasoning from perception to context evaluation to action justification, which CoT can explicitly represent during classification. Secondly, the variability in how explanations can be interpreted is enhanced by the aggregation mechanism in SC, which is robust to individual model predictions. In particular, we utilise three GPT model copies, each using CoT reasoning independently, and self consistency voting to predict the final context label.

\subsection{Corpus-Based Analysis of Domain-Specific Language}
\label{subsec:corpus_analysis}

Corpus linguistics offers a robust set of tools for examining patterns in vocabulary and grammar across specialised collections of text. These methods have been widely applied in professional and technical domains, consistently showing that word choice and sentence structure reflect the communicative context and functional demands of each field~\cite{brezina2018statistics, mcenery2011corpus}.

\subsubsection{Lexical Features in Specialised Corpora}

Identifying unique vocabulary patterns has proven essential in understanding how language is shaped by domain-specific requirements. Brookes and Collins~\cite{brookes2023corpus} present a comprehensive approach to applying corpus methods in health communication, showing how keyness analysis highlights systematic lexical distinctions between expert-led and patient-oriented discourse. By comparing how words are used in different areas or types of language, researchers have learned more about how language changes depending on the situation or context. Brezina~\cite{brezina2018statistics} developed rigorous statistical techniques for comparing lexical features across corpora, while McEnery and Brookes~\cite{mcenery2024corpus} have shown how such methods can uncover sociolinguistic trends in large-scale datasets. Collectively, these studies confirm that lexical choices are shaped by situational demands rather than being random or idiosyncratic, offering solid empirical foundations for analysing domain-specific language.

\subsubsection{Syntactic Patterns in Instructional and Explanatory Language}

Syntactic analysis is particularly insightful for understanding how language is used to instruct and explain. Studies of procedural texts have identified recurring clause structures and argument patterns that encode sequences of actions and causal links~\cite{faghihi2023role}. Research on tutorial dialogues suggests that syntactic complexity is closely tied to pedagogical goals, which means simpler structures tend to be used for direct instruction, while more complex ones serve elaboration and explanation~\cite{graesser2016conversations}. This reinforces the idea that syntax is functionally linked to communicative intent. In the field of human-robot interaction, researchers have explored how people naturally phrase commands and explanations when communicating with autonomous systems. Liu et al.~\cite{liu2024review} provide an extensive review of natural-language instruction systems for robotic execution, illustrating how users adopt consistent grammatical patterns when issuing commands or seeking clarification. Studies on navigation instructions have identified preferred syntactic forms for expressing spatial relationships and movement intentions~\cite{kollar2014grounding}. Bärmann et al.~\cite{barmann2024incremental} highlight how language used in robot interactions follows predictable structural patterns, which can support incremental learning. Together, these findings suggest that domain-specific instructional language is highly structured and may inform the development of more intuitive explanation systems.

\subsubsection{A Gap in the Analysis of Driving-Related Explanations}

Despite the extensive literature on lexical and syntactic patterns in specialised corpora, there has been limited exploration of how people linguistically construct explanations for driving decisions. Existing datasets with textual annotations, such as BDD-X~\cite{kim2018textual}, have primarily been used to train machine learning models for generating explanations, rather than as subjects of detailed linguistic analysis. Previous corpus-based work in the transportation domain has mostly focused on crash reports~\cite{chen2022traffic} or legal and regulatory language~\cite{chowdhury2023applications}, with little attention paid to real-time explanatory discourse from drivers or annotators. This gap is important because a better understanding of how people naturally explain driving actions can directly inform the design of explainable AV systems. If such explanations consistently follow certain lexical and syntactic patterns based on driving context, these patterns could be leveraged to generate system outputs that feel more familiar and understandable to users. This research addresses this overlooked area by applying corpus-based methods to analyse how human explanations are structured across different driving scenarios, offering an evidence-based foundation for generating scenario-aware AV explanations.
\section{Methodology}

\subsection{Dataset}

This study utilised a pre-processed version of the BDD-X dataset~\cite{kim2018textual}, previously employed in RAG Driver~\cite{yuan2024ragdriver}. The BDD-X dataset was originally developed to support research on explainable automated driving systems, with the goal of training models capable of generating natural language explanations for vehicle control decisions. It extends the Berkeley DeepDrive (BDD) dataset~\cite{yu2020bdd100k}, which comprises approximately 40-second dashboard-camera videos collected from human-driven vehicles operating under diverse environmental and traffic conditions. BDD-X has become a benchmark dataset in the field of explainable AVs and has been adopted in numerous studies investigating vision--language models for driving explanation and action prediction. Its widespread use can be attributed to several characteristics that make it particularly well suited for explanation research: (1) The explanations were authored by trained human annotators adopting a driving instructor’s perspective, recruited via Amazon Mechanical Turk from a pool of qualified workers who passed a test and demonstrated familiarity with U.S. driving rules, ensuring pedagogically meaningful justifications; (2) Annotations explicitly link observable driving actions to their underlying rationale; and (3) The dataset captures a wide range of real-world driving scenarios. We selected BDD-X for the current study because its human-authored explanations provide a realistic proxy for the type of context-rich driving commentary. In this study, only the textual explanations produced by human annotators were used. The original BDD-X dataset separates each explanation into \emph{action} and \emph{justification} components. However, we employed a merged version that combines both fields to represent complete, context-rich explanations of driving contexts. In total, the dataset comprised 16{,}392 explanation instances. Each explanation contained an average of 13.8 words (median = 13), indicating concise yet semantically complete utterances typical of driving-instruction commentary.


\subsection{Context Classification of Explanations}
\begin{figure}[t]
    \centering
    \includegraphics[width=\textwidth]{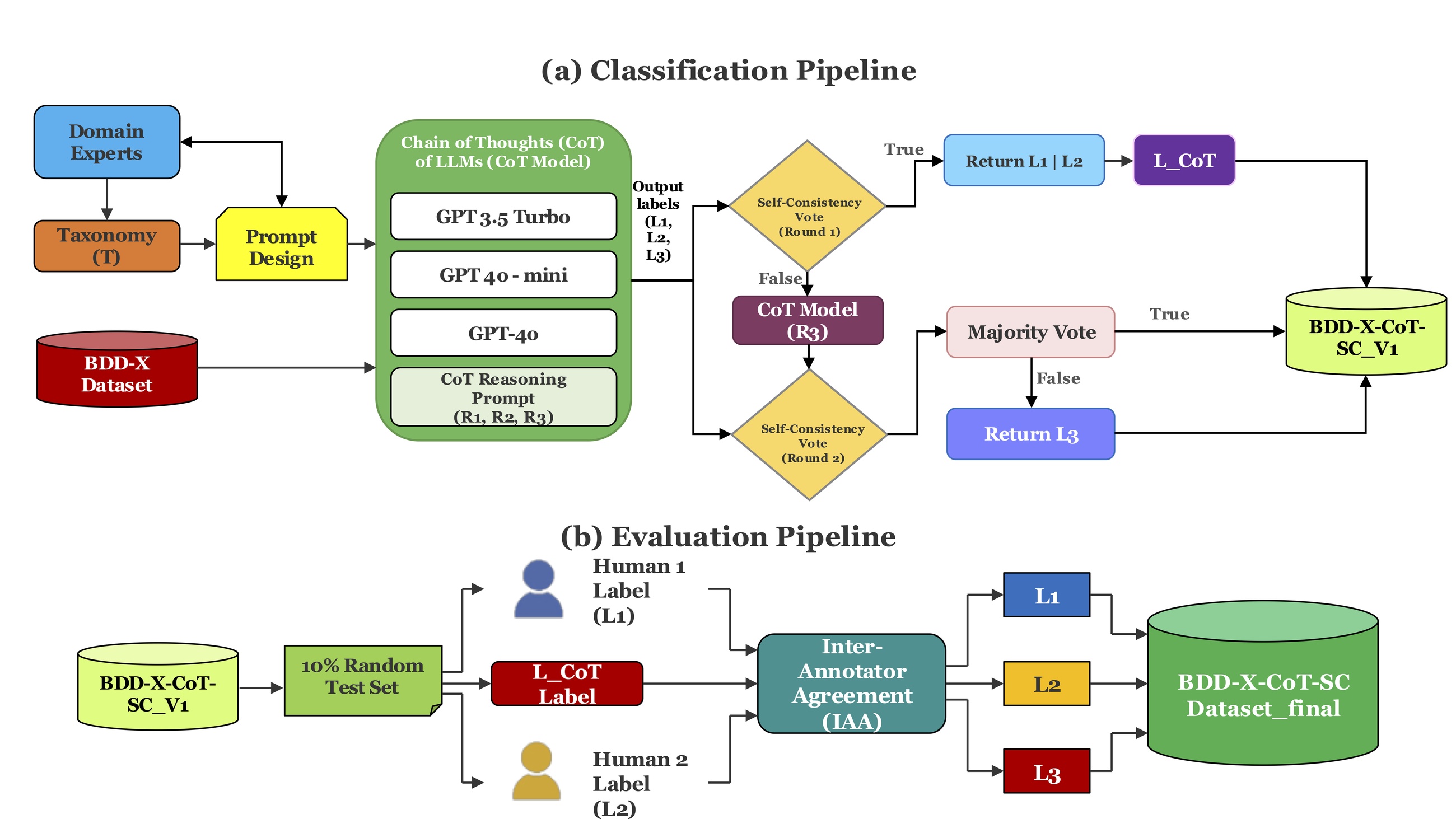}
    \caption{X-Blocks Level 1: The proposed RACE framework, (a) a CoT-SC-based classification pipeline for context scenario labelling of explanations in driving and (b) an evaluation pipeline for assessing reliability and human-model agreement.}
    \label{fig:xblocks_level1}
\end{figure}

The first level of the X-Blocks framework addresses the context block. To address this classification task, we propose the Reasoning-Aligned Classification of Explanations (RACE) framework. Figure~\ref{fig:xblocks_level1} illustrates the overall structure of RACE, which comprises two main pipelines: (a) the classification pipeline and (b) the evaluation pipeline.
The classification pipeline employs a CoT-SC based reasoning strategy across multiple LLMs to generate, validate, and consolidate scenario labels for explanations in the BDD-X dataset.
The evaluation pipeline then compares these model-generated labels against two independent human annotators (one with expertise in AI and AVs and the other with a background in human factors and transportation) to quantify inter-annotator agreement (IAA) and evaluate the reliability of the RACE model's reasoning process. Together, these components establish an interpretable, reasoning-aligned methodology for large-scale driving-explanation context classification, providing the foundational context labels upon which subsequent lexical and syntactic analyses depend.

\subsubsection{Driving Scenario Taxonomy}
The taxonomy was developed through an iterative, multi-stage process to ensure both theoretical grounding and empirical validity.
\textit{Step 1:} An initial list of driving context scenarios was compiled, drawing upon established frameworks and prior literature in scenario categorisation for the assessment of automated driving \cite{degelder2020scenario,baby2024development,taxonomy2017}.
\textit{Step 2:} A transportation research expert reviewed this preliminary taxonomy by systematically reviewing a wide range of publicly available dash-cam recordings and incorporating insights from over ten years of driving experience.
\textit{Step 3:} The taxonomy then underwent multiple rounds of collaborative review and refinement within the authorship team. During these sessions, the definitions, boundaries, and hierarchical relationships between scenario categories were iteratively revised to enhance conceptual clarity, internal consistency, and practical applicability.

A domain-specific taxonomy is iteratively refined through expert consensus rather than algorithmic optimisation:
\begin{equation}
T^{(0)} \xrightarrow{\text{initial}}
T^{(1)} \xrightarrow{\text{expert review}} \cdots
\xrightarrow{\text{expert review}} T^{*}
\end{equation}

where each taxonomy version $T^{(k)} = \{(t_i, d_i)\}_{i=1}^{K}$ consists of a finite set of $K$ label--definition pairs, with $t_i$ denoting the $i$-th taxonomy label and $d_i$ its corresponding natural-language definition. The process continues until domain experts reach a stable consensus, producing the final taxonomy $T^{*}$, which serves as the semantic reference for all subsequent classification stages. A detailed illustration of the proposed taxonomy can be found in Appendix~\ref{app:taxonomy}.

\subsubsection{Prompt Design and Reasoning}
Let $T^{*} = \{(t_k, d_k)\}_{k=1}^{K}$ denote a fixed taxonomy of $K = 32$ categories, where each category is defined by a label name $t_k$ and a textual definition $d_k$.
Let $D_s = \{(x_i, L_i^{\mathrm{true}})\}_{i=1}^{n}$ be a small development set of size $n \in [10,20]$, where $x_i$ is a natural-language explanation and $L_i^{\mathrm{true}} \in T^{*}$ is its ground-truth label.
The prompt template at refinement iteration $j$ is denoted by $Q^{(j)}$, with $Q^{(0)} = f_{\mathrm{prompt}}(T^{*}, D_s)$ obtained via an initial manual prompt-construction procedure.
Applying $Q^{(j)}$ to each $x_i$ yields predicted labels $\hat{L}^{(j)}_i \in T^{*}$, from which the empirical accuracy is computed as:
\begin{equation}
\mathrm{Acc}^{(j)} = \frac{1}{n} \sum_{i=1}^{n} \mathbb{I}\!\left[\hat{L}^{(j)}_i = L_i^{\mathrm{true}}\right],
\end{equation}
where $\mathbb{I}[\cdot]$ denotes the indicator function.
If $\mathrm{Acc}^{(j)}$ exceeds a predefined acceptance threshold $\tau_{\mathrm{accept}}$, the prompt is accepted as $Q^{*}$.
Otherwise, qualitative error patterns $\mathcal{E}^{(j)}$ are identified and used by a manual refinement operator $\textsc{ManualRefine}(\cdot)$ to produce the next prompt $Q^{(j+1)}$.

\begin{algorithm}[t]
\caption{Prompt design and manual refinement}
\label{alg:prompt_design}
\footnotesize
\begin{algorithmic}[1]
\Require Taxonomy $T^{*}$; development set $D_s=\{(x_i,L_i^{\mathrm{true}})\}_{i=1}^{n}$; acceptance threshold $\tau_{\mathrm{accept}}$
\Ensure Final prompt template $Q^{*}$

\State $Q^{(0)} \gets f_{\mathrm{prompt}}(T^{*},D_s)$ \Comment{Manual initial design}

\For{$j = 0,1,2,\ldots$}
    \State \textbf{(a) Apply prompt and predict labels}
    \For{$i = 1$ \textbf{to} $n$}
        \State $\hat{L}^{(j)}_i \gets \mathrm{InferLabel}(Q^{(j)},T^{*},x_i)$
    \EndFor

    \State \textbf{(b) Compute empirical accuracy}
    \State $\mathrm{Acc}^{(j)} \gets \frac{1}{n} \sum_{i=1}^{n}
    \mathbb{I}\!\left[\hat{L}^{(j)}_i = L_i^{\mathrm{true}}\right]$

    \If{$\mathrm{Acc}^{(j)} \ge \tau_{\mathrm{accept}}$}
        \State \Return $Q^{*} \gets Q^{(j)}$
    \EndIf

    \State \textbf{(c) Qualitative error analysis and refinement}
    \State $\mathcal{E}^{(j)} \gets \mathrm{ErrorAnalysis}
    \!\left(\{(x_i,\hat{L}^{(j)}_i,L_i^{\mathrm{true}})\}_{i=1}^{n}\right)$
    \State $Q^{(j+1)} \gets \mathrm{ManualRefine}(Q^{(j)},\mathcal{E}^{(j)})$
\EndFor
\end{algorithmic}
\end{algorithm}

\subsubsection{CoT Annotation with Self-Consistency}

We next apply the final prompt $Q^{*}$, obtained from Algorithm~\ref{alg:prompt_design}, to the full unlabelled dataset $D=\{x_i\}_{i=1}^{N}$ using a self-consistent ensemble of Chain of Thought (CoT) enabled language models.

Given the optimised prompt template $Q^{*}$ and taxonomy $T^{*}$, each input explanation $x_i \in D$ is annotated using an ensemble $\mathcal{M}=\{M_1,M_2,M_3\}$, comprising \texttt{gpt-3.5-turbo}, \texttt{gpt-4o-mini}, and \texttt{gpt-4o}. A critical design decision is the exclusive use of models from the GPT family rather than combining models from different families. From a theoretical perspective, self-consistency \cite{wang2022self} derives its effectiveness from sampling diverse reasoning paths within a shared model family, rather than from architectural heterogeneity. Leveraging capacity variation across GPT models induces diverse reasoning trajectories while preserving the theoretical foundations of self-consistency.

From a practical perspective, single-family ensembles ensure semantic commensurability, calibration consistency, and reproducibility. Models within the GPT family share tokenisation schemes, training methodologies, and alignment conventions~\cite{brown2020language,achiam2023gpt}, ensuring directly comparable outputs and reliable majority voting. Prior work has shown that cross-family ensembles can introduce systematic biases due to divergent output distributions and response formatting conventions~\cite{jiang2023llm}. Our approach thus maximises reasoning-path diversity while maintaining controlled conditions for reliable self-consistency aggregation.

For each input $x_i$, a taxonomy-conditioned prompt $Q_{x_i}=Q^{*}(x_i,T^{*})$ is instantiated, and each model $M_i$ produces a reasoning trace $R_i$. Each trace is projected onto the taxonomy via the projection operator
\begin{equation}
\pi_{T^{*}}(R_i,T^{*})=\arg\max_{(t_k,d_k)\in T^{*}} s(R_i,d_k),
\end{equation}
where $s(\cdot,\cdot)$ denotes a semantic similarity function.

\begin{algorithm}[t][htbp]
\caption{CoT annotation with self-consistency (conditional tiebreak)}
\label{alg:cot_sc}
\footnotesize
\begin{algorithmic}[1]
\Require Taxonomy $T^{*}$; final prompt template $Q^{*}$; dataset $D=\{x_i\}_{i=1}^{N}$; base models $M_1,M_2$ and tiebreak model $M_3$
\Ensure For each $x_i$: $(x_i,L^{*},\{R_i\}_{i=1}^{3},\hat{p}(L^{*}))$

\ForAll{$x_i \in D$}
    \State $Q_{x_i} \gets Q^{*}(x_i,T^{*})$

    \State \textbf{Generate reasoning traces from base models}
    \State $R_1 \gets M_1^{\mathrm{CoT}}(Q_{x_i},T^{*})$
    \State $R_2 \gets M_2^{\mathrm{CoT}}(Q_{x_i},T^{*})$

    \State \textbf{Project each trace to a taxonomy label}
    \State $L_1 \gets \pi_{T^{*}}(R_1,T^{*})$
    \State $L_2 \gets \pi_{T^{*}}(R_2,T^{*})$

    \State \textbf{Round 1: two-model consensus}
    \If{$L_1 = L_2$}
        \State $L^{*} \gets L_1$
        \State $R_3 \gets \emptyset$
    \Else
        \State \textbf{Round 2: conditional tiebreak}
        \State $R_3 \gets M_3^{\mathrm{CoT}}(Q_{x_i},T^{*})$
        \State $L_3 \gets \pi_{T^{*}}(R_3,T^{*})$

        \State $L^{\mathrm{maj}} \gets
        \arg\max_{\ell \in T^{*}} \sum_{i=1}^{3} \mathbb{I}[L_i=\ell]$

        \If{$\sum_{i=1}^{3} \mathbb{I}[L_i=L^{\mathrm{maj}}] \ge 2$}
            \State $L^{*} \gets L^{\mathrm{maj}}$
        \Else
            \State $L^{*} \gets L_3$
        \EndIf
    \EndIf

    \State $\hat{p}(L^{*}) \gets
    \mathrm{AggregateConfidence}(L^{*},\{L_i\}_{i=1}^{3},\{R_i\}_{i=1}^{3})$

    \State \textbf{Output:} $(x_i,L^{*},\{R_i\}_{i=1}^{3},\hat{p}(L^{*}))$
\EndFor
\end{algorithmic}
\end{algorithm}

\subsubsection{Evaluation}
\label{subsec:evaluation}

We evaluate the reliability and validity of the proposed reasoning-based classification framework using a multi-dimensional protocol that combines classification performance, inter-annotator agreement, and explanation consistency. The evaluation addresses three complementary aspects: (i) correctness of predicted driving-context labels, (ii) agreement between model outputs and human annotations, and (iii) consistency among multiple annotators.

To assess annotation reliability, we constructed a gold-standard evaluation subset from the BDD-X explanation corpus. A stratified random sample comprising 10\% of the data was drawn while preserving the empirical distribution of driving-context labels. Only labels with at least two instances were included in the stratified sampling pool; all underrepresented labels ($n_i < 2$) were retained in full to ensure coverage of rare categories. Formally, for taxonomy labels $t_i \in T^{*}$ with $n_i \ge 2$, the sampled set is defined as
\[
D_{\mathrm{sample}} = \mathrm{StratifiedSample}(D,\,0.10 \mid T^{*}),
\]
and the final gold-standard set as:
\[
D_{\mathrm{gold}} = D_{\mathrm{sample}} \cup D_{\mathrm{rare}}.
\]
This procedure yields a balanced and reproducible evaluation subset.

Two independent domain experts (Human~1 and Human~2) annotated all instances in $D_{\mathrm{gold}}$ using Label Studio. Annotators worked independently and were blinded to both model predictions and each other’s labels, following a shared annotation guideline. Instances where Human~1 and Human~2 agreed were recorded as \emph{H1\&H2 agreement}. Disagreements were resolved through a consensus meeting, producing a reconciled \emph{Human Consensus} label.

Model performance and annotation reliability were quantified using four complementary metrics: accuracy (Acc), macro-averaged F$_1$ ($\mathrm{F}_1^{\mathrm{macro}}$), Cohen’s $\kappa$~\cite{cohen1960coefficient}, and Fleiss’ $\kappa_F$~\cite{fleiss1971measuring}. Accuracy measures the proportion of correct predictions, while macro-F$_1$ captures class-balanced performance as:
\begin{equation}
\mathrm{F}_1^{\mathrm{macro}} = \frac{1}{K} \sum_{k=1}^{K} \frac{2 P_k R_k}{P_k + R_k},
\end{equation}
where $P_k = TP_k/(TP_k + FP_k)$ and $R_k = TP_k/(TP_k + FN_k)$ denote precision and recall for class $k$, with $TP_k$, $FP_k$, and $FN_k$ representing true positives, false positives, and false negatives.

Pairwise agreement between model predictions and human annotations was assessed using Cohen’s $\kappa$, computed as:
\begin{equation}
\kappa = \frac{p_o - p_e}{1 - p_e},
\end{equation}
where $p_o$ and $p_e$ represent observed and expected agreement, respectively. Cohen’s $\kappa$ was computed for model vs.\ Human~1, model vs.\ Human~2, model vs.\ Human Consensus, and model vs.\ the H1\&H2 agreement subset. Agreement strength was interpreted using the Landis--Koch scale (0.41--0.60 = moderate, 0.61--0.80 = substantial, 0.81--1.00 = almost perfect)~\cite{landis1977measurement}.

To assess agreement among multiple raters, Fleiss’ $\kappa_F$ was computed over the set $\{\text{Human~1}, \text{Human~2}, \text{Model}\}$ as:
\begin{equation}
\kappa_F = \frac{\bar{P} - \bar{P}_e}{1 - \bar{P}_e},
\end{equation}
where $\bar{P}$ denotes mean observed agreement and $\bar{P}_e$ is the mean expected agreement across all labels. Together, Acc and $\mathrm{F}_1^{\mathrm{macro}}$ quantify classification correctness and balance across driving contexts, Cohen’s $\kappa$ captures pairwise human–model alignment, and Fleiss’ $\kappa_F$ provides a holistic measure of multi-rater agreement. This combination yields a robust evaluation of both predictive accuracy and explanatory consistency.


\subsection{Syntax Block in Driving Explanation}
\label{subsec:syntax_blocks}

\begin{figure}[t]
\centering
\includegraphics[width=\textwidth]{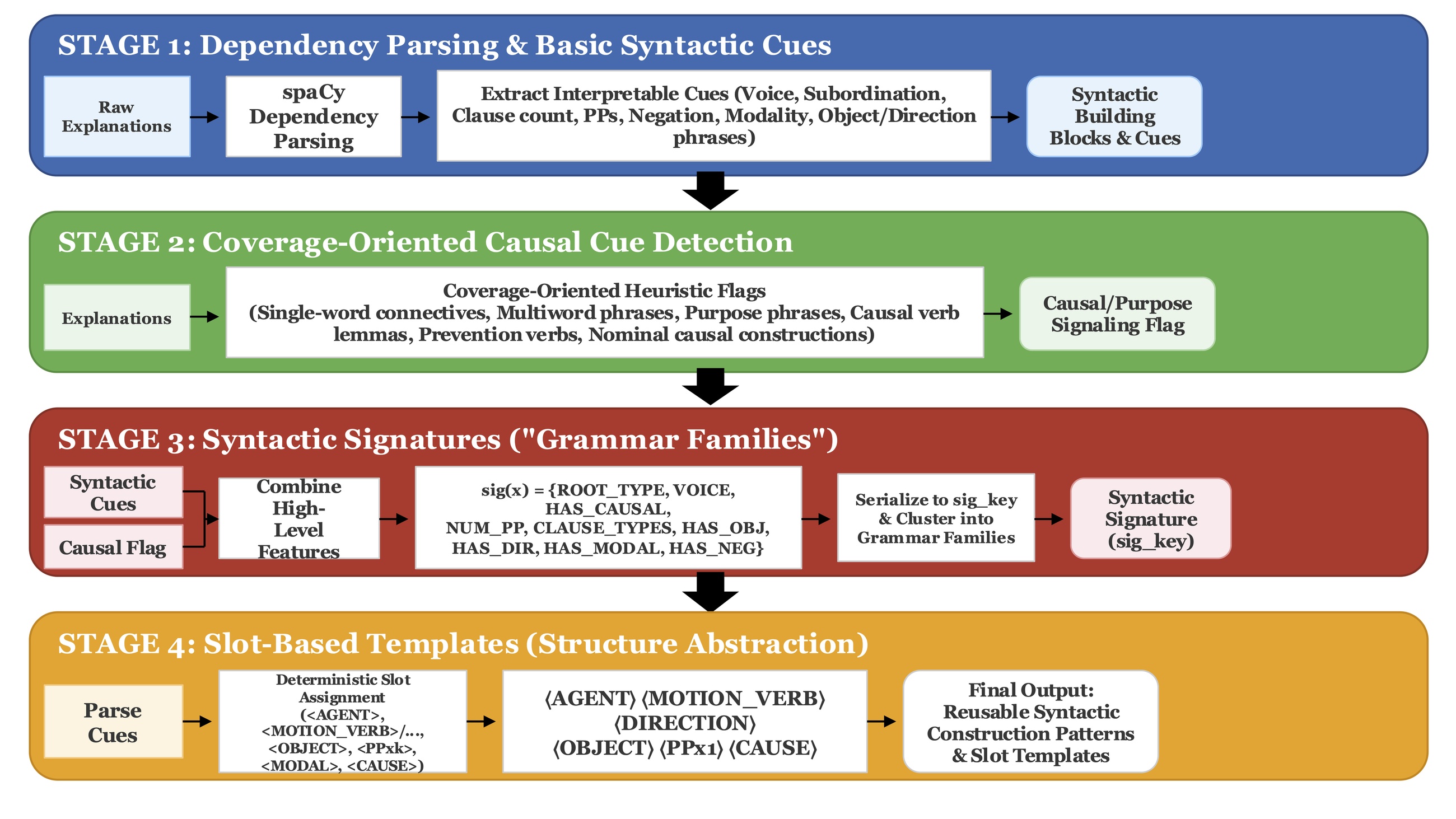}
\caption{X-Blocks Level~2: Extracting syntactic building blocks of driving explanations. Raw explanations are parsed using dependency parsing to obtain interpretable syntactic cues (Stage~1), augmented with coverage-oriented detection of causal and purpose constructions (Stage~2), combined into discrete syntactic signatures representing grammar families (Stage~3), and abstracted into reusable slot-based templates that preserve high-level structural patterns (Stage~4).}
\label{fig:syntactic}
\end{figure}

The second level of the X-Blocks framework addresses the \emph{Syntax} block. This level captures how explanations are linguistically structured by analysing sentence-level syntax and grammatical relations, abstracting away from lexical content while preserving interpretable structural patterns. Figure \ref{fig:syntactic} illustrates the overall pipeline of the syntax block.

\subsubsection{Dependency Parsing and Units of Analysis}
Each explanation is parsed using a dependency parser, producing a directed tree in which nodes correspond to tokens and edges correspond to grammatical relations. We adopt a dependency-based representation following the Universal Dependencies formalism~\cite{nivre2016universal}, which provides a consistent and cross-domain inventory of syntactic relations. Rather than operating on full parse trees, we focus on a compact set of interpretable, linguistically motivated features derived from dependency relations that are directly relevant to explanation structure.

\subsubsection{Clause Structure and Grammatical Features}
From each dependency parse, we extract a set of coarse syntactic features that characterise sentence-level structure. Specifically, we consider:
(i) \emph{grammatical voice}, classified as active or passive based on the presence of passive subjects and auxiliary constructions;
(ii) \emph{subordinate clause presence}, detected via dependency relations corresponding to adverbial, complement, and relative clauses;
(iii) \emph{number of clauses}, approximated as one main clause plus the number of subordinate clause heads; and
(iv) \emph{prepositional phrase count}, defined as the number of tokens participating in prepositional dependency relations.
These features follow standard dependency-based analyses of clause structure~\cite{de2008stanford}. In addition, binary indicators capture the presence of negation, modal auxiliaries, explicit subjects and objects, and directional phrases. Together, these features provide a stable abstraction of syntactic form across diverse surface realisations of driving explanations.

\subsubsection{Causal and Purpose Constructions}
Because driving explanations often encode rationales for actions and decisions, we explicitly model causal structure. An explanation is marked as containing a causal or purpose construction if it satisfies any of the following conditions: the presence of causal discourse markers (e.g., \textit{because}, \textit{therefore}); multiword causal expressions (e.g., \textit{due to}, \textit{as a result}); purpose phrases (e.g., \textit{in order to}); causal or preventive verb lemmas (e.g., \textit{cause}, \textit{avoid}); or nominal causal constructions realised through prepositional attachments. This rule-based detection strategy is inspired by prior work on discourse connectives and the automatic identification of causal relations~\cite{prasad2017penn,girju2003automatic}.

\subsubsection{Syntactic Signatures}
To identify recurring structural patterns, we define a \emph{syntactic signature} as a discrete vector of syntactic attributes extracted from an explanation. Each signature encodes the root verb type (motion, stop/yield, state, or other), grammatical voice, presence of causal marking, number of prepositional phrases, clause configuration, and selected phrase-level indicators. Explanations sharing the same syntactic signature are grouped into \emph{grammar families}, which represent reusable syntactic building blocks underlying explanation structure.

\subsubsection{Slot-Based Templates}
To further abstract syntactic form, we derive slot-based templates that replace lexical content with functional placeholders while preserving grammatical structure. Slots include \texttt{<AGENT>}, \texttt{<MOTION\_VERB>}, \texttt{<STOP\_VERB>}, \texttt{<OBJECT>}, \texttt{<PP\_k>}, \texttt{<NEG>}, and \texttt{<CAUSE>}. These templates retain high-level argument structure and syntactic cues, enabling comparison of explanation forms across diverse driving scenarios while abstracting away surface-level lexical variation.

\subsubsection{Cross-Context Structural Reuse}
Finally, we analyse the distribution of syntactic signatures and slot-based templates across driving contexts. Structures that recur across multiple contexts indicate general-purpose explanation grammars, whereas those concentrated within a single context reveal specialised explanatory strategies tailored to specific driving situations.

\subsection{Morphemes and Lexemes Block in Driving Explanation}
\label{subsec:morphemes_lexemes}

\begin{figure}[t]
\centering
\includegraphics[width=\textwidth]{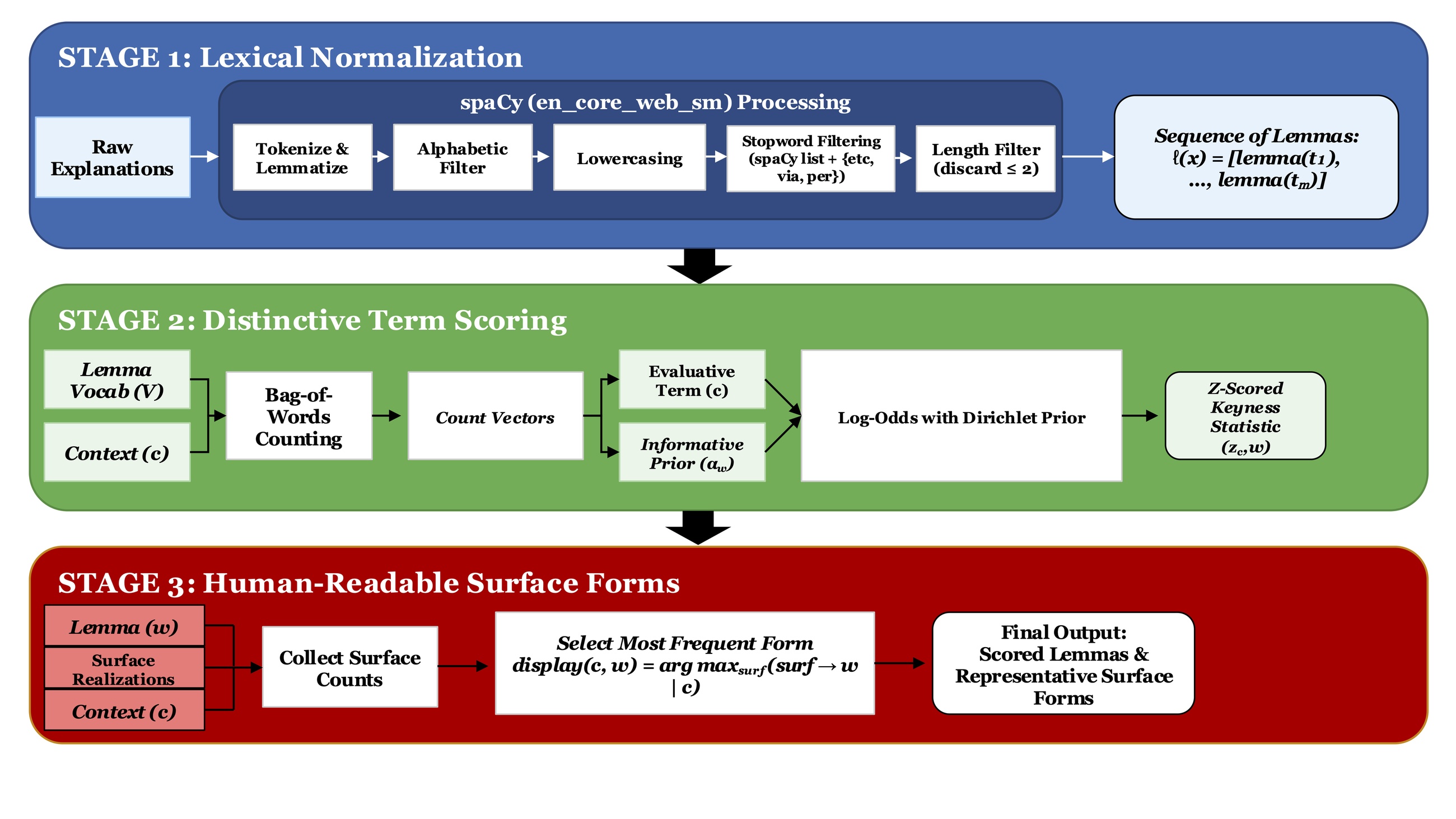}
\caption{X-Blocks Level~3: Pipeline for extracting morpheme- and lexeme-level building blocks of driving explanations. Raw explanations are normalised via lemmatisation and filtering (Stage~1), scored for contextual distinctiveness using log-odds with an informative Dirichlet prior (Stage~2), and mapped back to representative surface forms for human interpretability (Stage~3).}
\label{fig:lexical}
\end{figure}

The third level of the X-Blocks framework addresses the \emph{Morphemes and Lexemes} block. This level captures the lexical building blocks of driving explanations at the level of morphemes and lexemes, with the objective of identifying content-bearing lexical units that characterise explanations within and across driving contexts. Figure \ref{fig:lexical} illustrates the overall pipeline of the lexical block.

\subsubsection{Text Units and Preprocessing}
An explanation is modelled as an ordered sequence of tokens $T = (t_1, t_2, \dots, t_n)$, where each token $t_i$ is obtained through standard tokenisation. Each explanation is processed independently using the spaCy \texttt{en\_core\_web\_sm} pipeline~\cite{honnibal2020spacy}, which performs tokenisation, lemmatisation, and part-of-speech (POS) tagging. Only alphabetic tokens are retained and normalised to lowercase.

A \emph{lemma} is defined as the canonical dictionary form of a token and is denoted by $\ell(t_i)$. Inflectional variants (e.g., \textit{cross}, \textit{crossing}, \textit{crossed}) are mapped to the same lemma. Stopwords are high-frequency function words that carry limited semantic content (e.g., \textit{at}, \textit{to}, \textit{an})and are removed using spaCy’s English stopword list augmented with a small domain-neutral set. Tokens of length less than or equal to two characters are excluded. The resulting normalised lemma sequence for an explanation is therefore
\[
L = (\ell_1, \ell_2, \dots, \ell_m),
\]
where $\ell_i = \ell(t_i)$ for each retained token.

\subsubsection{Lexical Frequency Representation}
For each driving context $c$, all explanations assigned to $c$ are aggregated to form a bag-of-lemmas representation. Let $n_{c,w}$ denote the number of occurrences of lemma $w$ in context $c$, and let $n_{\neg c,w}$ denote the number of occurrences of $w$ in the complement set consisting of all other contexts. We use $\neg c$ to denote the complement of context $c$ (one-vs.-rest), i.e., the set of explanations assigned to contexts other than $c$.

The total number of lemma tokens in context $c$ is
\begin{equation}
N_c = \sum_{w \in \mathcal{V}} n_{c,w},
\end{equation}
where $\mathcal{V}$ denotes the lemma vocabulary. Similarly, the total number of lemma tokens in the complement set is
\begin{equation}
N_{\neg c} = \sum_{w \in \mathcal{V}} n_{\neg c,w}.
\end{equation}
This representation transforms labelled explanations into context-specific and background lexical counts, enabling principled comparison of lexical usage across driving scenarios.

\subsubsection{Lexical Keyness via Log-Odds with Informative Prior}
\label{subsec:lexical_keyness}

To measure how strongly a lemma is associated with a particular driving context, we compute a lexical \emph{keyness} score using the log-odds ratio with an informative Dirichlet prior. Originally proposed by Monroe et al.~\cite{monroe2008fightin}, this method yields robust estimates even when contexts differ substantially in size or when many lemmas are rare. Related approaches have been widely adopted in corpus linguistics and natural language processing to identify context-specific lexical patterns~\cite{gries2016quantitative,eisenstein2011discovering}.

Each lemma $w$ is assigned a background-based pseudo-count $\alpha_w$, reflecting its frequency in the full corpus. This prior encodes a weak expectation about typical lexical usage and mitigates spurious effects arising from sparse data. Let $\alpha_0$ denote the total prior mass over the vocabulary $\mathcal{V}$:
\begin{equation}
\alpha_0 = \sum_{w \in \mathcal{V}} \alpha_w
\label{eq:alpha_sum}
\end{equation}

The log-odds difference for lemma $w$ between context $c$ and the background distribution is then computed as:
\begin{equation}
\delta_{c,w}
=
\log\frac{n_{c,w} + \alpha_w}{N_c + \alpha_0 - (n_{c,w} + \alpha_w)}
-
\log\frac{n_{\neg c,w} + \alpha_w}{N_{\neg c} + \alpha_0 - (n_{\neg c,w} + \alpha_w)}
\label{eq:log_odds_difference}
\end{equation}

To enable comparison across lemmas with differing frequencies, this quantity is normalised by its estimated uncertainty. Following~\cite{monroe2008fightin}, the variance of the log-odds estimate yields the standardised $z$-score:
\begin{equation}
z_{c,w}
=
\frac{\delta_{c,w}}
{\sqrt{\frac{1}{n_{c,w} + \alpha_w} + \frac{1}{n_{\neg c,w} + \alpha_w}}}
\label{eq:keyness_zscore}
\end{equation}

Large positive values of $z_{c,w}$ indicate that lemma $w$ is disproportionately associated with context $c$, whereas negative values indicate relative underuse. This formulation highlights meaningful lexical building blocks while reducing noise from rare terms and small sample sizes.

\subsubsection{Surface Form Reconstruction}
Although lexical keyness is computed at the lemma level, results are reported using representative surface forms to improve interpretability. For each context–lemma pair $(c,w)$, we select the most frequently observed surface realisation of $w$ in explanations associated with context $c$. This mapping preserves linguistic naturalness in reporting while maintaining statistical robustness through lemma-based scoring. Overall, this morpheme- and lexeme-level analysis captures the content-bearing lexical units that form the foundational building blocks of natural language explanations in driving scenarios.


\section{Results}
\label{sec:results}

\subsection{Context Classification Results}
\label{subsec:results_context_classification}

This section reports the empirical results of the proposed RACE framework for classifying natural-language driving explanations into scenario-aware context categories. Results are presented in terms of label distribution, agreement with human annotators, and class-wise prediction behaviour.

\begin{table}[t]
\centering
\caption{Inter-rater agreement between models and human annotators. Metrics include accuracy (Acc), macro-averaged F$_1$ ($\mathrm{F}_1^{\mathrm{macro}}$), Cohen’s $\kappa$, Fleiss’ $\kappa$, and qualitative agreement level.}
\label{tab:agreement_multilabel_final}
\resizebox{\textwidth}{!}{
\begin{tabular}{l l c c c c l}
\toprule
\textbf{Model} & \textbf{Compared With} & \textbf{Acc (\%)} & $\boldsymbol{\mathrm{F}_1^{\mathrm{macro}}}$ & \textbf{Cohen’s $\kappa$} & \textbf{Fleiss’ $\kappa$} & \textbf{Agreement Level} \\
\midrule
\textbf{RACE (CoT-SC)} & Human 1 (H1) & 90.37 & 0.85 & 0.89 & 0.83 & \textbf{Almost Perfect} \\
 & Human 2 (H2) & 79.88 & 0.51 & 0.76 & 0.83 & Substantial \\
 & Human Consensus & 60.22 & 0.47 & 0.57 & 0.83 & Moderate \\
 & H1 \& H2 & \textbf{91.45} & \textbf{0.88} & \textbf{0.91} & \textbf{0.83} & \textbf{Almost Perfect} \\
\midrule
\textbf{GPT-3.5-Turbo (CoT)} & Human 1 (H1) & 77.44 & 0.62 & 0.74 & 0.74 & Substantial \\
 & Human 2 (H2) & 70.24 & 0.43 & 0.65 & 0.74 & Substantial \\
 & Human Consensus & 50.56 & 0.36 & 0.46 & 0.74 & Moderate \\
 & H1 \& H2 & 79.63 & 0.66 & 0.76 & 0.74 & Substantial \\
\midrule
\textbf{GPT-4o-mini (CoT)} & Human 1 (H1) & 80.18 & 0.65 & 0.77 & 0.76 & Substantial \\
 & Human 2 (H2) & 72.56 & 0.47 & 0.67 & 0.76 & Substantial \\
 & Human Consensus & 48.70 & 0.33 & 0.45 & 0.76 & Moderate \\
 & H1 \& H2 & 82.13 & 0.70 & 0.79 & 0.76 & Substantial \\
\midrule
\textbf{GPT-4o (CoT)} & Human 1 (H1) & 82.44 & 0.77 & 0.80 & 0.72 & Substantial \\
 & Human 2 (H2) & 66.00 & 0.33 & 0.61 & 0.72 & Moderate \\
 & Human Consensus & 54.55 & 0.43 & 0.51 & 0.72 & Moderate \\
 & H1 \& H2 & \textbf{84.22} & \textbf{0.82} & \textbf{0.82} & \textbf{0.72} & \textbf{Almost Perfect} \\
\bottomrule
\end{tabular}}
\end{table}

\subsubsection{Distribution of Driving Context Labels}
\begin{figure}[t]
\centering
\includegraphics[width=\textwidth]{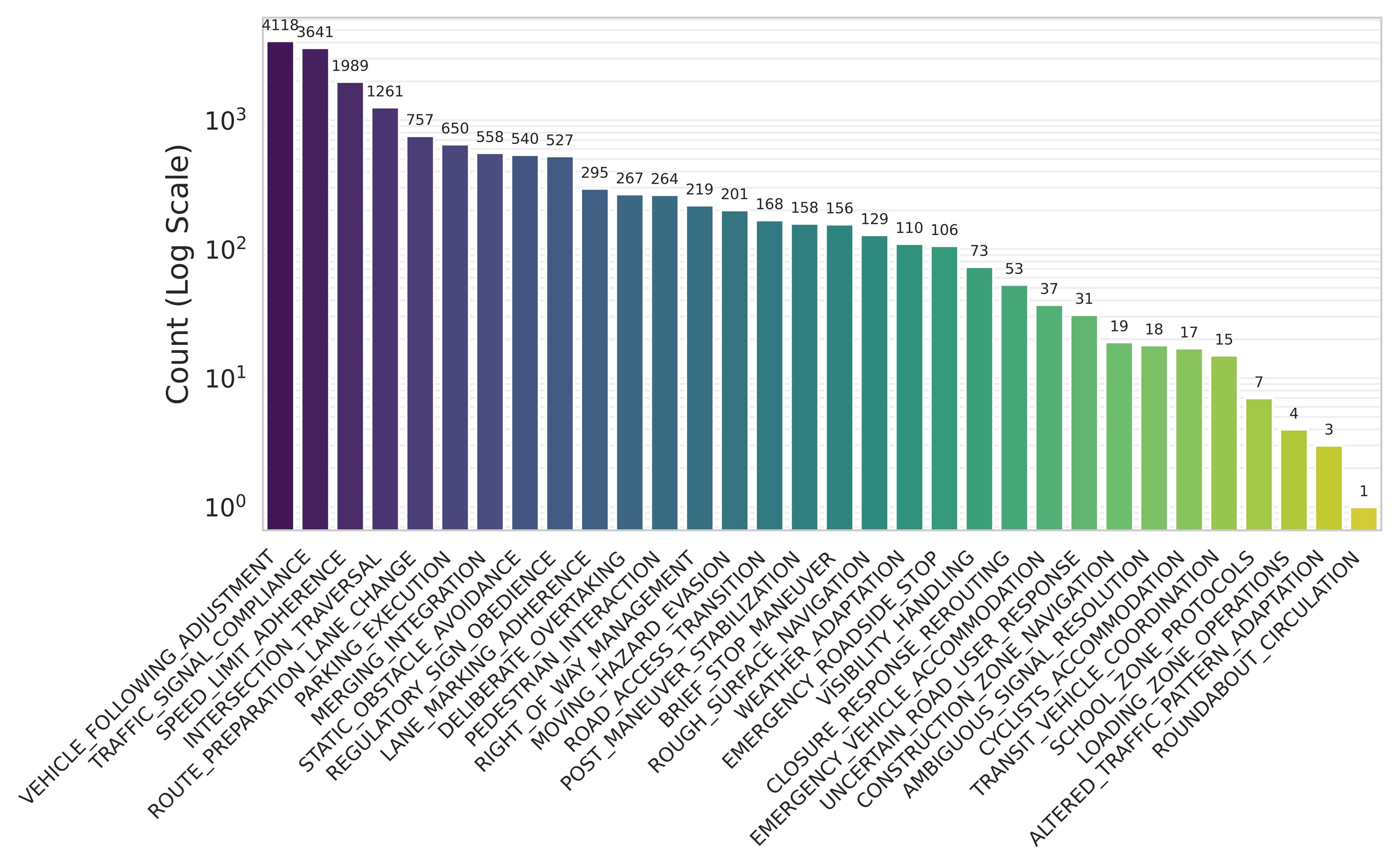}
\caption{Distribution of driving-context labels in the annotated BDD-X explanation corpus. Counts are shown on a logarithmic scale, revealing a highly imbalanced, long-tailed distribution across scenario contexts.}
\label{fig:context_dist}
\end{figure}

Figure~\ref{fig:context_dist} illustrates the distribution of driving-context labels assigned to explanations in the BDD-X corpus. The frequency of labels follows a highly skewed, long-tailed distribution spanning several orders of magnitude. High-frequency contexts such as \textit{Vehicle Following Adjustment}, \textit{Traffic Signal Compliance}, and \textit{Speed Limit Adherence} account for a substantial portion of the dataset, while several contexts occur only rarely, including \textit{Roundabout Circulation}, \textit{Closure Operations}, and \textit{Altered Traffic Pattern Adaptation}. This distribution reflects the composition of the BDD-X dataset, which was originally collected for training and evaluating video-to-text explanation models~\cite{kim2018textual} rather than to provide balanced coverage across all possible driving contexts. The observed imbalance motivates the use of agreement-based metrics and macro-averaged performance measures to ensure that evaluation is not dominated by high-frequency categories.

\begin{figure}[t]
\centering
\includegraphics[width=\textwidth]{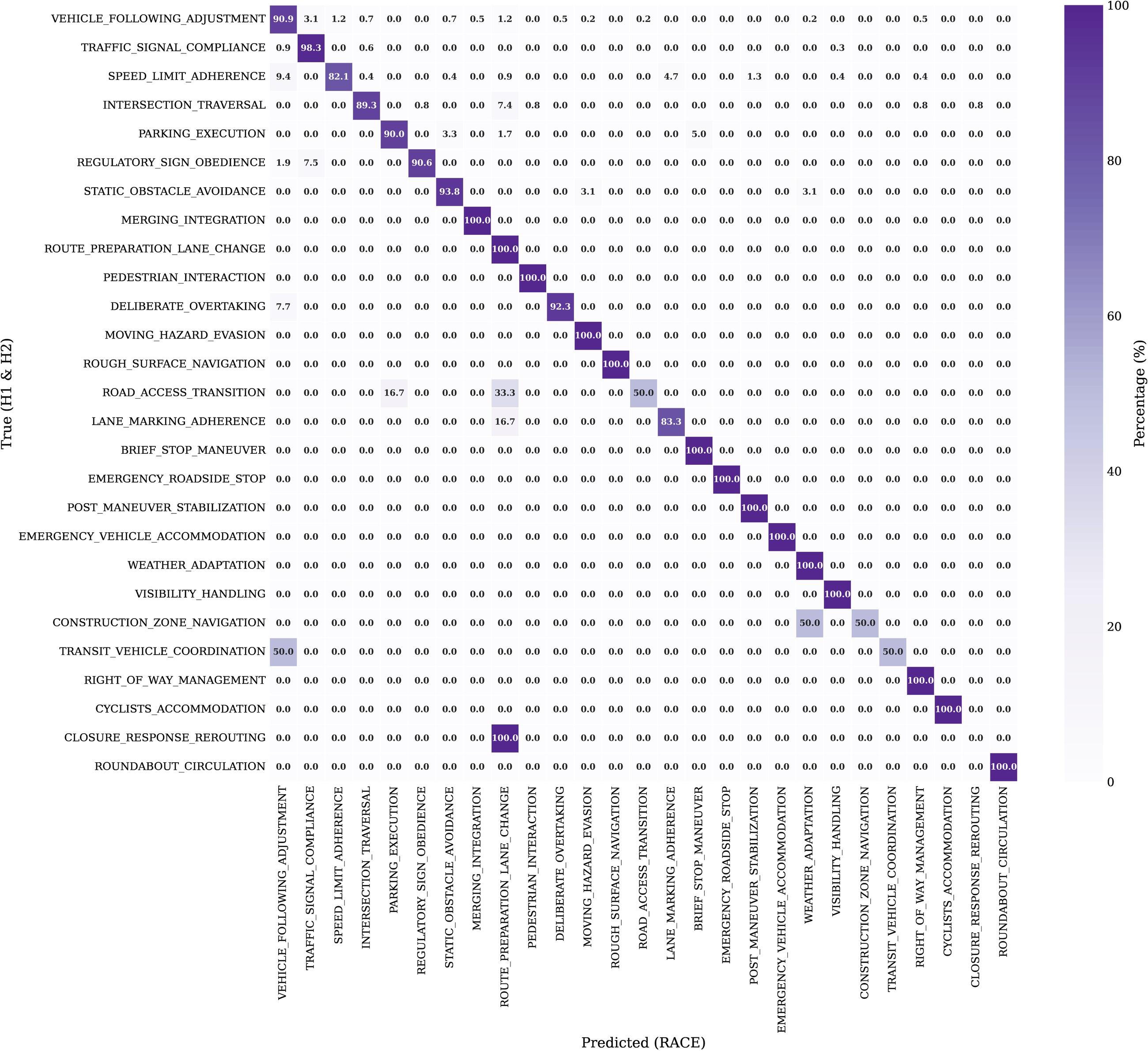}
\caption{Normalised confusion matrix (\% per true label) comparing RACE predictions with the H1\&H2 human agreement labels. Strong diagonal dominance indicates robust class-wise agreement, with limited confusion among semantically related driving contexts.}
\label{fig:confusion_matrix}
\end{figure}

\subsubsection{Human--Model Agreement and Classification Performance}
Table~\ref{tab:agreement_multilabel_final} reports agreement between model predictions and human annotations using accuracy (Acc), macro-averaged F$_1$ ($\mathrm{F}_1^{\mathrm{macro}}$), Cohen’s $\kappa$, Fleiss’ $\kappa$, and qualitative agreement levels. RACE achieves high agreement with individual human annotators, reaching 90.37\% accuracy and a Cohen’s $\kappa$ of 0.89 when compared with Human~1, and 79.88\% accuracy with $\kappa = 0.76$ when compared with Human~2. Agreement with the subset of explanations where both annotators agree (H1\&H2) reaches 91.45\% accuracy and $\kappa = 0.91$, corresponding to an \emph{Almost Perfect} agreement level. Across all evaluation settings, RACE consistently outperforms single-model CoT baselines in both accuracy and agreement metrics. Fleiss’ $\kappa$ values further indicate strong multi-rater consistency among the model and human annotators.

\subsubsection{Class-wise Prediction Behaviour}
Figure~\ref{fig:confusion_matrix} presents the normalised confusion matrix comparing RACE predictions against the H1\&H2 human agreement labels. Strong diagonal dominance is observed across most driving contexts, indicating high class-wise agreement. Confusion is limited and primarily occurs between semantically related contexts, such as \textit{Route Preparation / Lane Change} and \textit{Merging Integration}, or \textit{Brief Stop Manoeuvre} and \textit{Emergency Roadside Stop}. Rare contexts exhibit increased variability, reflecting limited sample sizes and higher annotation uncertainty. Overall, these results demonstrate that RACE achieves robust and consistent classification of natural-language driving explanations across a diverse and imbalanced set of scenario contexts.

\subsection{Lexical Building Blocks Results Across Driving Contexts}
\label{subsec:lexical_results}

This section reports the results of the lexical keyness analysis described in Section~\ref{subsec:lexical_keyness}. Using log-odds ratios with an informative Dirichlet prior, we identify lemmas that are statistically overrepresented in explanations associated with each driving context relative to all other contexts.

\begin{figure}[tbp]
\centering
\includegraphics[width=\textwidth]{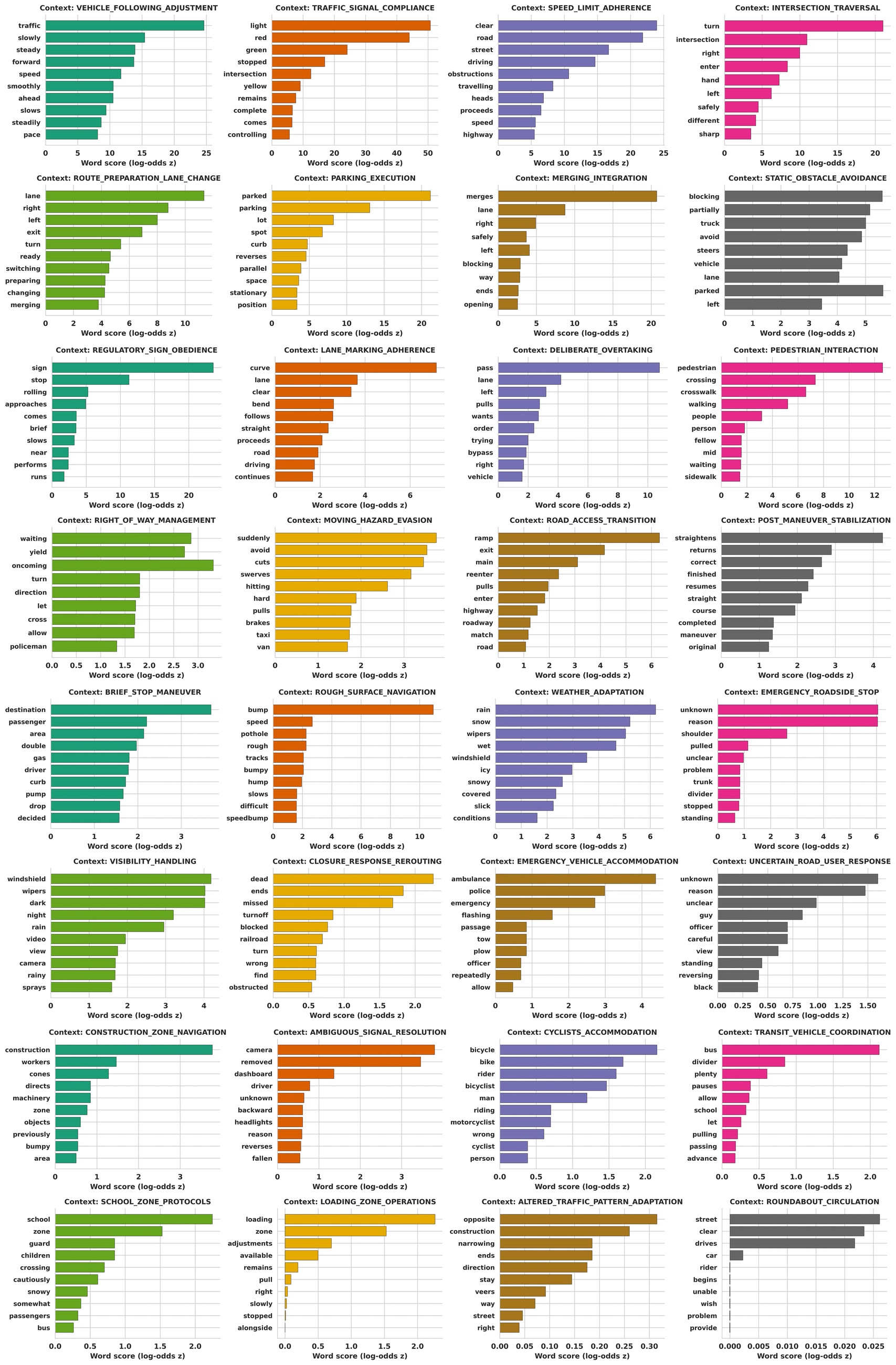}
\caption{Distinctive lexical building blocks across driving contexts. Each panel shows the top lexical markers for a subset of scenarios, ranked by log-odds keyness ($z$-score). Terms are displayed using representative surface forms for interpretability.}
\label{fig:lexeme_keyness}
\end{figure}

Figure~\ref{fig:lexeme_keyness} visualises the top-ranked lemmas for each of the 32 driving contexts, with bars indicating the standardised log-odds ($z$) score for each lemma. Higher scores correspond to stronger contextual association. Importantly, these lexemes do not define or constitute driving contexts. Rather, they serve as distinctive linguistic markers that characterise how humans verbally describe and explain actions within each context. The lexical building blocks identified here therefore reveal vocabulary patterns that differentiate explanations across contexts, providing insight into the semantic focus of human explanatory language.

Across contexts, we observe clear and semantically coherent lexical signatures. For example, explanations classified under \textit{Vehicle Following Adjustment} are characterised by motion-regulation terms such as \textit{traffic}, \textit{slowly}, \textit{steady}, and \textit{speed}, reflecting a focus on longitudinal control and pacing when humans describe car-following behaviour. In contrast, \textit{Traffic Signal Compliance} explanations exhibit strong associations with signal-state terminology, including \textit{light}, \textit{red}, \textit{green}, and \textit{stopped}, indicating that humans emphasise traffic control devices when explaining signal-related actions. Speed-related contexts such as \textit{Speed Limit Adherence} prominently feature lemmas like \textit{clear}, \textit{road}, \textit{street}, and \textit{driving}, suggesting that explanations in these contexts are framed around roadway conditions and lawful progression. Intersection-centric contexts, including \textit{Intersection Traversal} and \textit{Route Preparation Lane Change}, show distinctive directional and spatial vocabulary such as \textit{turn}, \textit{left}, \textit{right}, \textit{lane}, and \textit{exit}.

Contexts involving interaction with other road users display strong entity-based lexical cues. For instance, \textit{Pedestrian Interaction} explanations are dominated by \textit{pedestrian}, \textit{crossing}, \textit{crosswalk}, and \textit{walking}, while \textit{Cyclist Accommodation} emphasises \textit{bicycle}, \textit{cyclist}, \textit{rider}, and \textit{bike}. Similarly, \textit{Emergency Vehicle Accommodation} shows high keyness for \textit{ambulance}, \textit{police}, \textit{emergency}, and \textit{flashing}. Environmental adaptation contexts reveal distinct lexical patterns tied to external conditions: \textit{Weather Adaptation} and \textit{Visibility Handling} are associated with terms such as \textit{rain}, \textit{snow}, \textit{windshield}, \textit{wipers}, and \textit{dark}. Road-surface-related contexts such as \textit{Rough Surface Navigation} highlight lemmas including \textit{bump}, \textit{pothole}, \textit{rough}, and \textit{speed bump}. Less frequent or inherently ambiguous contexts, such as \textit{Uncertain Road User Response} and \textit{Ambiguous Signal Resolution}, exhibit lower-magnitude $z$-scores and more generic lexical items (e.g., \textit{unknown}, \textit{reason}, \textit{unclear}), reflecting both reduced data support and semantic uncertainty.

Overall, these results demonstrate that the proposed lexical keyness method successfully extracts interpretable, context-sensitive lexical markers that reveal how humans linguistically differentiate their explanations across driving scenarios. These lexical building blocks do not define what a context \emph{is}, but instead capture the distinctive vocabulary humans employ when explaining actions within each context.

\subsection{Syntactic Building Blocks Results Across Driving Contexts}
\label{subsec:syntax_results}

\begin{figure}[tbp]
    \centering
    \includegraphics[width=\textwidth]{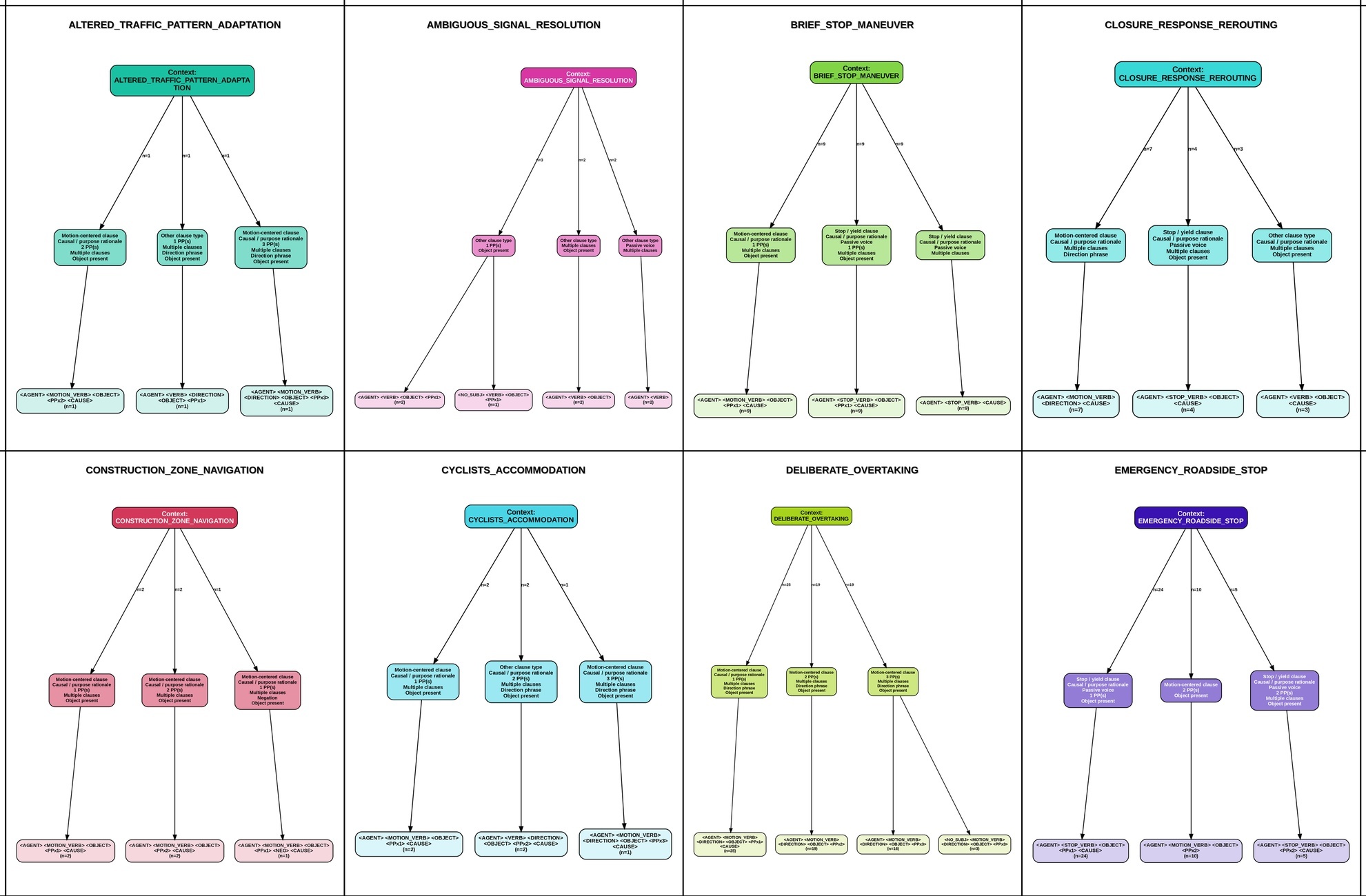}\par
    \vspace{2mm}
    \includegraphics[width=\textwidth]{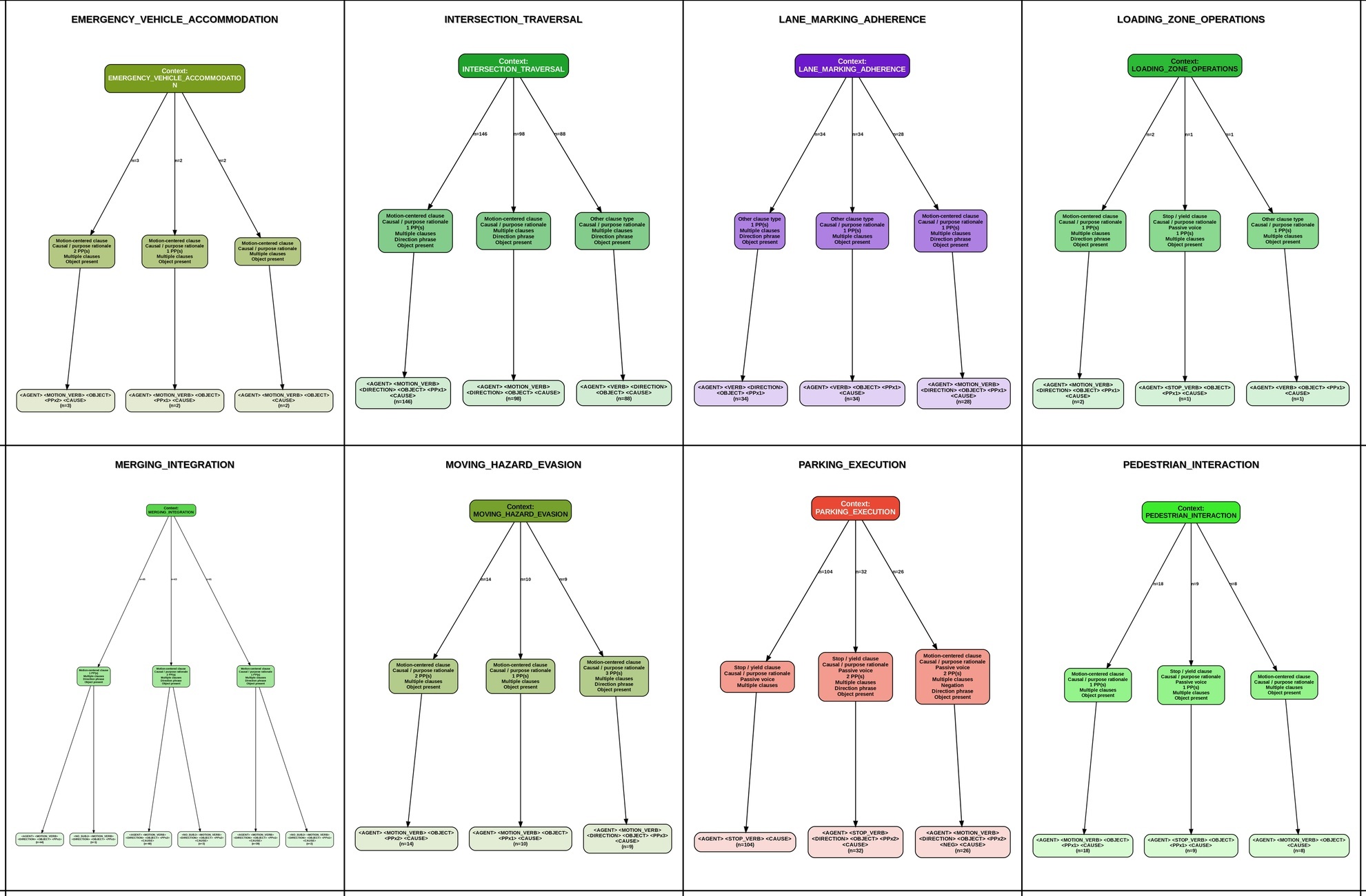}
    \caption{Syntactic grammar hierarchies of driving explanations (Part 1). For each driving context, explanations are grouped into grammar families based on shared syntactic signatures and instantiated as slot-based templates.}
    \label{fig:syntax1}
\end{figure}

\begin{figure}[tbp]
    \centering
    \includegraphics[width=\textwidth]{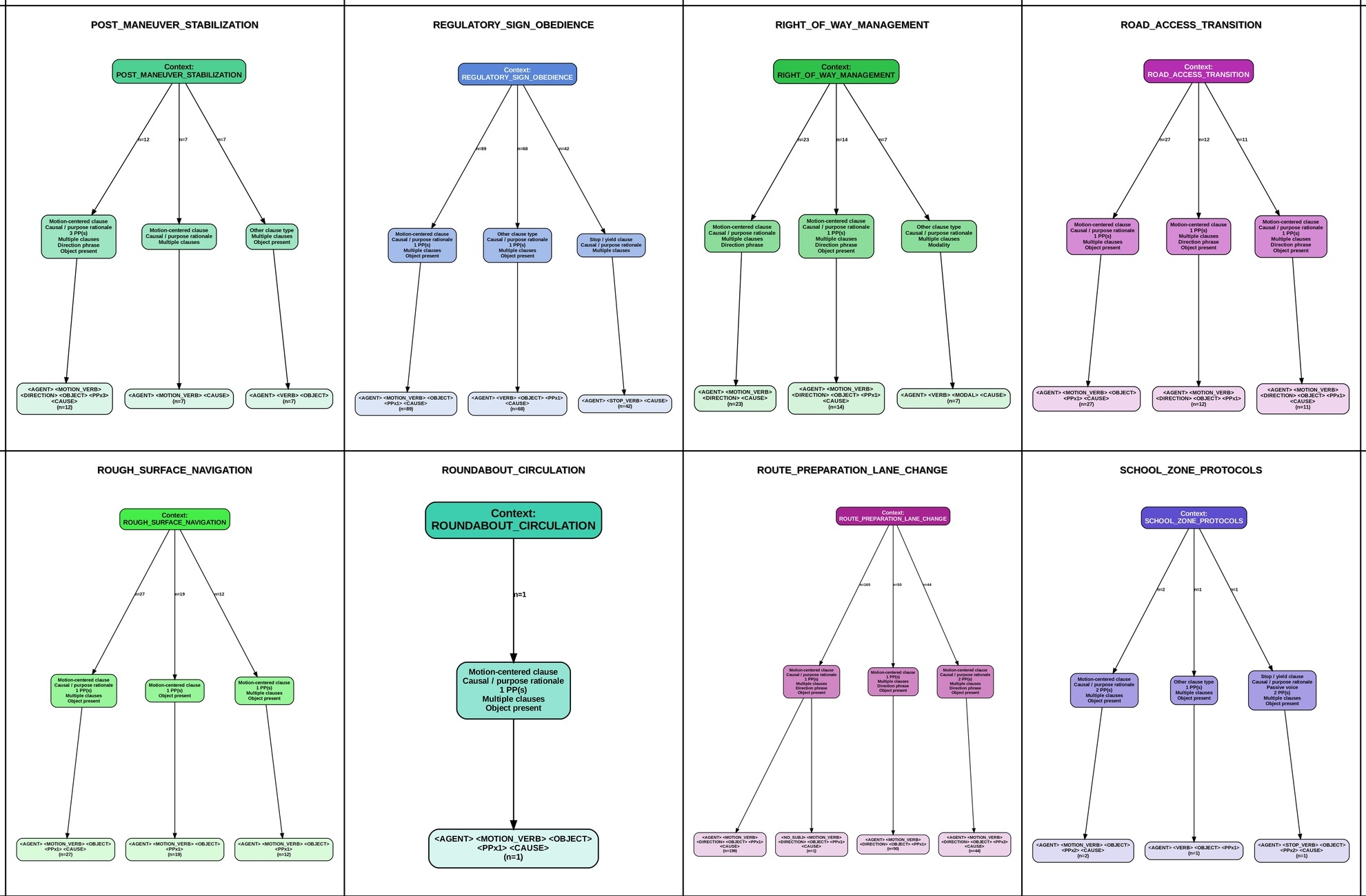}\par
    \vspace{2mm}
    \includegraphics[width=\textwidth]{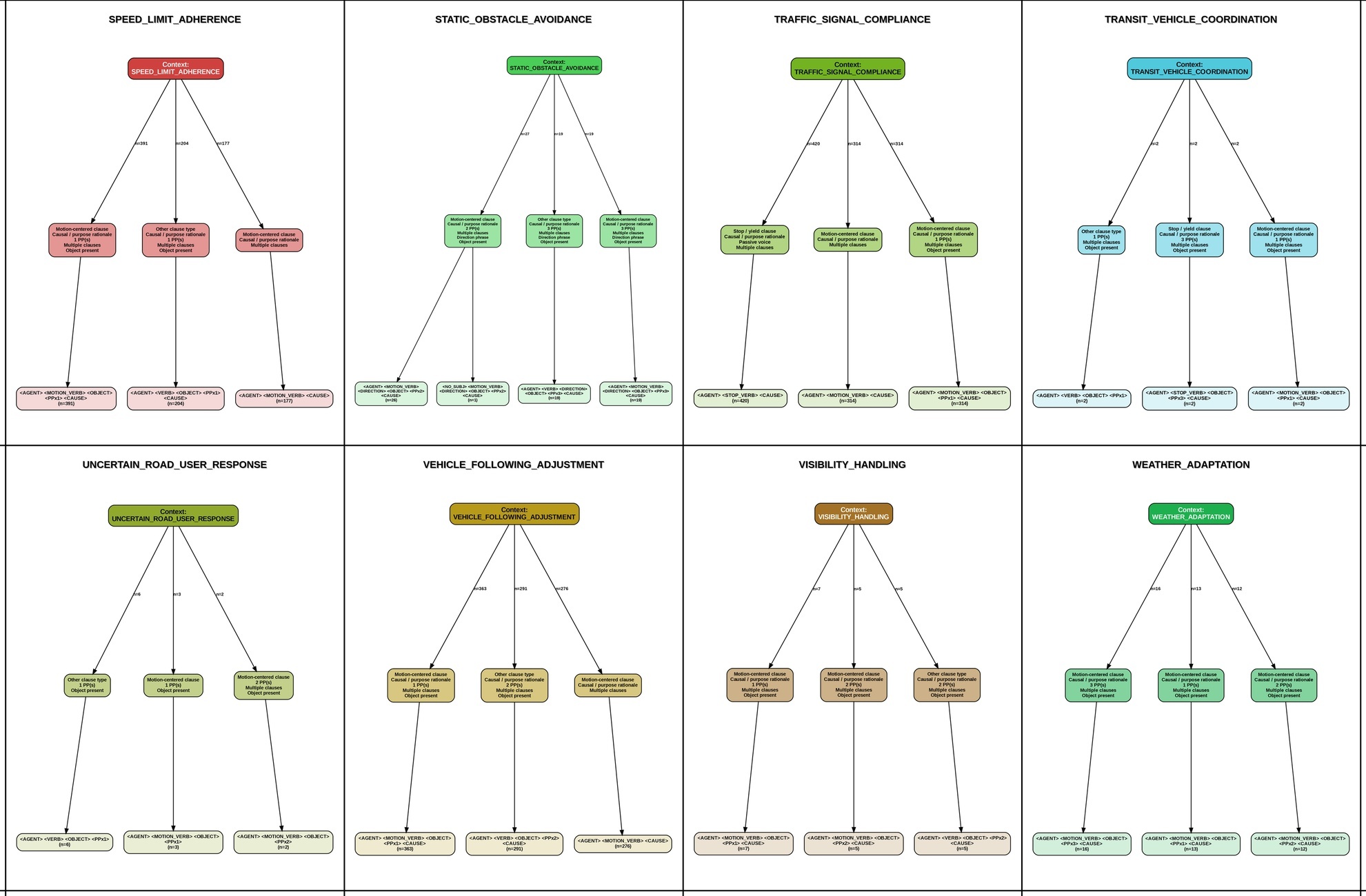}
    \caption{Syntactic grammar hierarchies of driving explanations (Part 2).}
    \label{fig:syntax2}
\end{figure}

This section reports the results of the syntactic building block analysis described in Section~\ref{subsec:syntax_blocks}. Using dependency parses, causal/purpose cue detection, syntactic signatures (``grammar families''), and slot-based templates, we group explanations into a small set of reusable structural patterns for each driving context. Figures~\ref{fig:syntax1} and~\ref{fig:syntax2} summarise the dominant grammar families and their representative templates across contexts. Across the analysed contexts, most explanations concentrate into a limited set of syntactic signatures.

In motion-centric contexts (e.g., \textit{Intersection Traversal}, \textit{Merging Integration}), dominant families are organised around a motion predicate with explicit spatial arguments, typically realised as an \texttt{<AGENT>} plus \texttt{<MOTION\_VERB>} followed by directional or locative modifiers. These families repeatedly instantiate templates of the form:
\[
\texttt{<AGENT> <MOTION\_VERB> <DIRECTION/LOCATION> <PP> [<CAUSE>]}.
\]
This pattern reflects that navigation explanations often foreground where the vehicle moved and why. Contexts centred on rules and compliance (e.g., \textit{Speed Limit Adherence}, \textit{Regulatory Sign Obedience}, \textit{School Zone Protocols}) are dominated by stop/state verb families and compact clause structures. Representative templates include:
\[
\texttt{<AGENT> <STOP\_VERB> <PP> [<CAUSE>]}
\quad \text{and} \quad
\texttt{<AGENT> <STATE\_VERB> [<CAUSE>]}.
\]
Compared with motion contexts, these families emphasise obligation and justification (e.g., stopping/obeying) rather than path description. Causal and purpose constructions are common across contexts, yet their placement differs. Planning-heavy contexts (e.g., \textit{Route Preparation Lane Change}, \textit{Deliberate Overtaking}) more frequently show multi-phrase expansions (multiple PPs and occasional subordination), consistent with anticipatory explanations. In contrast, reactive safety contexts (e.g., \textit{Static Obstacle Avoidance}, \textit{Moving Hazard Evasion}) typically realise justification in short single-clause forms, indicating immediate rationale rather than extended planning.

Several high-level templates recur across many contexts, especially structures that combine (i) an explicit agent, (ii) a main action predicate, and (iii) an optional causal clause or PP rationale:
\[
\texttt{<AGENT> <VERB> [<OBJECT/DIRECTION>] [<CAUSE>]}.
\]
This reuse suggests that human driving explanations draw from a shared syntactic repertoire, while context identity is often expressed by (a) the predicate type (motion vs.\ stop/state) and (b) the presence/shape of spatial and causal modifiers. Table~\ref{tab:template_examples} provides illustrative examples for each template type, showing how slot-based patterns map to natural language explanations.

\begin{table}[tbp]
\centering
\caption{Illustrative examples of syntactic templates with slot-to-text mappings. Subscripts indicate slot assignments: AGT = Agent, MOT\_V = Motion Verb, STP\_V = Stop Verb, STA\_V = State Verb, DIR = Direction, PP = Prepositional Phrase, CAU = Cause.}
\label{tab:template_examples}
\small
\setlength{\tabcolsep}{6pt}
\renewcommand{\arraystretch}{1.15}
\begin{tabularx}{\linewidth}{>{\raggedright\arraybackslash}p{0.33\linewidth} >{\raggedright\arraybackslash}X}
\toprule
\textbf{Template} & \textbf{Example with Slot Mapping} \\
\midrule

\textbf{\texttt{<AGT> <MOT\_V> <DIR> <PP> [<CAU>]}} &
\textit{[The car]$_{\mathrm{AGT}}$ [turns]$_{\mathrm{MOT\_V}}$ [left]$_{\mathrm{DIR}}$ [at the intersection]$_{\mathrm{PP}}$ [because the light is green]$_{\mathrm{CAU}}$} \\
\addlinespace[2pt]

\textbf{\texttt{<AGT> <STP\_V> <PP>}} &
\textit{[The car]$_{\mathrm{AGT}}$ [stops]$_{\mathrm{STP\_V}}$ [at the red light]$_{\mathrm{PP}}$} \\
\addlinespace[2pt]

\textbf{\texttt{<AGT> <STA\_V> [<CAU>]}} &
\textit{[The car]$_{\mathrm{AGT}}$ [is waiting]$_{\mathrm{STA\_V}}$ [for pedestrians to cross]$_{\mathrm{CAU}}$} \\
\addlinespace[2pt]

\textbf{\texttt{<AGT> <VERB> [<OBJ/DIR>] [<CAU>]}} &
\textit{[The car]$_{\mathrm{AGT}}$ [slows down]$_{\mathrm{VERB}}$ [to let the pedestrian cross]$_{\mathrm{CAU}}$} \\
\bottomrule
\end{tabularx}
\end{table}

\section{Discussion}
\label{sec:discussion}

The introduced X-Blocks framework is a hierarchical analytical approach that decomposes natural language driving explanations into three linguistic levels: context, syntax, and lexicon. At the context level, the RACE pipeline classifies explanations into scenario categories, providing the foundation for subsequent lexical and syntactic analyses. The following subsections discuss the findings at each level in relation to prior work and their implications for explainable automated vehicle design. To the best of our knowledge, existing studies have not provided a unified analysis of human-authored driving explanations that simultaneously considers contextual, lexical, and syntactic dimensions. Accordingly, rather than performing direct methodological comparisons, we situate our findings within established theoretical and empirical work in explainable AVs, corpus linguistics, and human–robot interaction.

\subsection{Context Classification Performance}
\label{subsec:discussion_context_performance}

The high agreement between RACE predictions and human annotations indicates that the proposed reasoning-aligned framework effectively captures the dominant contextual intent expressed in human-authored driving explanations. The achieved Cohen’s $\kappa$ of 0.91 against the H1\&H2 agreement subset places RACE in the \emph{almost perfect} agreement range according to the Landis–Koch scale~\cite{landis1977measurement}, and compares favourably with inter-annotator agreement levels reported in other complex NLP classification tasks. For example, recent annotation studies in specialised domains such as medical text classification and named entity recognition typically report $\kappa$ values between 0.70 and 0.86~\cite{artstein2017inter}, despite involving smaller label inventories. The strong agreement observed here is notable given the relatively large taxonomy of 32 driving-context categories.

This performance suggests that RACE aligns most strongly with cases where humans themselves exhibit high confidence and consistency. In particular, it supports the claim that the framework identifies salient explanatory cues rather than exploiting superficial lexical correlations, consistent with findings by Wei et al.~\cite{wei2022chain} showing that Chain of Thought prompting enables genuine multi-step reasoning. The consistent improvements over single-model CoT baselines further highlight the value of combining CoT reasoning with self-consistency and conditional tie-breaking. This result aligns with Wang et al.~\cite{wang2022self}, who demonstrated that self-consistency substantially improves reasoning accuracy by marginalising over diverse reasoning paths. While their work focused on arithmetic and commonsense reasoning, our findings extend these benefits to domain-specific text classification, suggesting that self-consistency mechanisms are broadly applicable to contextual inference tasks.

Residual confusion between closely related driving contexts reflects inherent ambiguities in natural language explanations rather than systematic model failure. Scenarios such as merging and route preparation often co-occur in real-world driving and may be described using overlapping linguistic cues. This observation is consistent with prior work on driving behaviour classification, which notes that real-world manoeuvres often share kinematic and contextual features~\cite{feng2023driving}. Similarly, distinctions between closely related contexts such as brief stops and emergency manoeuvres depend on subtle shifts in emphasis. Comparable ambiguity has been observed in temporal action segmentation, where annotators frequently disagree on action boundaries~\cite{ding2023temporal}. These confusion patterns therefore mirror known challenges in human annotation and suggest that strict one-to-one mappings between explanations and contexts may not always reflect human reasoning.

Beyond the BDD-X dataset, the RACE framework offers a generalisable methodology for classifying free-form natural language into domain-specific taxonomies. Its core components are taxonomy-conditioned prompting, Chain of Thought reasoning, and self-consistency, which are domain-agnostic and readily transferable. Recent applications of LLM-based annotation in medical question answering~\cite{singhal2023large} and legal document classification~\cite{kuznietsov2024explainable} suggest broad applicability. In AV industry settings, RACE can support scalable annotation of driver feedback, incident reports, and naturalistic driving narratives, and can serve as a prerequisite step for context-aware explanation generation systems.

\subsection{Lexical Patterns in Driving Explanations}
\label{subsec:lexical_discussion}

The lexical analysis reveals that human-authored driving explanations rely on highly systematic and context-sensitive word choices, supporting the premise that explanations are composed of reusable lexical building blocks aligned with situational demands. This finding aligns with corpus-linguistic research demonstrating that lexical choices in specialised domains are shaped by communicative requirements rather than being random or idiosyncratic~\cite{gries2021new,brezina2018statistics}. The observed lexical signatures indicate that explanations prioritise salient causal factors within each context. For example, traffic-signal contexts foreground signal states, while pedestrian-related contexts foreground social actors.

A clear distinction emerges between action-centric contexts (e.g., lane changes, overtaking, merging), which emphasise verbs and directional terms, and condition-centric contexts (e.g., weather adaptation, visibility handling), which emphasise environmental descriptors. This pattern mirrors findings in register variation research, where procedural texts prioritise action verbs while descriptive texts foreground noun phrases and adjectives~\cite{biber2019register}. These results suggest that humans adapt their explanatory register depending on whether justification is rooted in intentional manoeuvring or external constraints.

Some lexical items (e.g., \textit{road}, \textit{lane}, \textit{vehicle}) appear across multiple contexts with moderate keyness scores, forming a shared explanatory backbone for driving discourse. In contrast, highly context-specific lemmas (e.g., \textit{crosswalk}, \textit{ambulance}, \textit{pothole}) function as discriminative lexical anchors. This structure aligns with findings from domain-specific corpus studies, which show that specialised discourse typically consists of a general lexical core supplemented by situation-specific terminology~\cite{brezina2018statistics}. From an explainability perspective, these results suggest that effective AV explanations should mirror human strategies by emphasising context-defining entities and conditions rather than exhaustive detail.

\subsection{Syntactic Patterns and Structural Reuse}
\label{subsec:syntax_discussion}

The syntactic analysis demonstrates that human driving explanations are structurally economical. Within a given context, a small number of grammar families account for the majority of explanations, and across contexts, many families share common template skeletons. This concentration suggests that explanation structure is governed by a limited and reusable set of syntactic patterns rather than unconstrained variation. Similar observations have been reported in studies of syntactic complexity in domain-specific writing~\cite{lei2023large,lu2021relationship}.

\textit{Implication 1: Explanation generation can rely on a small template inventory.}
The consistent recurrence of dominant templates—typically agent–action structures with optional causal justification—indicates that human-like explanations do not require unconstrained syntactic generation. Instead, explanation systems can select from a limited, context-appropriate inventory of templates and populate them with situation-specific content. This approach aligns with template-based natural language generation methods, which offer improved controllability and auditability in safety-critical domains~\cite{gatt2018survey}.

\textit{Implication 2: Context identity is expressed through structural choices.}
Differences between driving scenarios are reflected primarily in predicate type and modifier structure rather than lexical choice alone. Motion-centric contexts favour action predicates with spatial arguments, whereas compliance-oriented contexts favour stop or state predicates. This observation is consistent with construction grammar theory, which posits that grammatical constructions themselves carry meaning independent of lexical content~\cite{hilpert2019construction}.

\textit{Implication 3: Causal reasoning is universal but context-sensitive.}
Causal justification appears across nearly all contexts, but its syntactic realisation varies with task demands. Planning-oriented scenarios often employ richer causal packaging, while reactive or safety-critical situations favour compact causal clauses. This finding supports design principles for explainable AVs: concise causal explanations are preferable in time-critical contexts, whereas more elaborate rationales are appropriate when future intent must be communicated~\cite{yousefizadeh2025psylingxav}.

Two factors contribute to observed structural variability within some contexts. First, some explanations compress multiple actions or intentions into a single sentence, producing multi-intent rationales. Second, dependency parsing of short, telegraphic utterances, which are common in driving narration can introduce ambiguity. These factors represent methodological limitations rather than properties of human explanatory behaviour and should be considered when interpreting grammar-family distributions.


\section{Limitations and Future Work}
\label{sec:limitations}

This study has several limitations that inform future research directions. First, the analysis relies exclusively on the English-language BDD-X dataset collected from U.S. driving scenarios. While the identified linguistic building blocks provide foundational insights, their generalisability to other languages and driving cultures remains to be established through cross-cultural validation studies. Second, the present analysis treats explanations as isolated textual units without integrating the temporal dynamics or multimodal context (video, sensor data) from which they originate. Future work should incorporate multimodal grounding to enable context detection directly from perceptual inputs and to investigate explanation timing, which is a factor identified as critical for user trust and comprehension. Finally, while X-Blocks provides foundational linguistic insights for explainable AV design, translating these building blocks into deployable real-time explanation systems requires further engineering development and user evaluation studies to assess their effectiveness in naturalistic driving settings.

\section{Conclusion}
\label{sec:conclusion}

This paper introduced X-Blocks, a framework that identifies the linguistic building blocks of natural language explanations for automated vehicles across three hierarchical levels: context, syntax, and lexicon. The proposed RACE framework, combining Chain of Thought reasoning with Self Consistency mechanisms across multiple LLMs, achieves 91.45\% accuracy and Cohen's $\kappa$ of 0.91 against human annotator agreement, demonstrating almost perfect alignment in classifying driving explanations into scenario-aware context categories. Lexical analysis reveals that human explanations employ context-specific vocabulary patterns, with action-centric scenarios emphasising motion verbs and directional terms while condition-centric scenarios foreground environmental descriptors. Syntactic analysis demonstrates that explanations concentrate into a limited repertoire of grammar families, with predicate type and causal construction complexity varying systematically across contexts. These findings indicate that human driving explanations are compositionally structured from interpretable, reusable linguistic units rather than arbitrary constructions. The X-Blocks framework provides actionable insights for designing scenario-aware explanation systems in AVs. The identified building blocks can inform template-based explanation generation that mirrors human communicative strategies, supporting the development of transparent and trustworthy AV communication that is both contextually appropriate and cognitively accessible to users.

\section*{Acknowledgements}
This research was supported by Queensland University of Technology (QUT) and the Australian Research Council Discovery Project funding scheme (DP220102598).

\section*{Data availability}
Data will be made available on request. The original BDD-X dataset is publicly available at \url{https://github.com/JinkyuKimUCB/BDD-X-dataset.git}.

\appendix

\clearpage
\section{Driving Context Taxonomy}
\label{app:taxonomy}

\begin{figure}[!htbp]
\centering
\includegraphics[width=1.2\textwidth,height=0.8\textheight,keepaspectratio]{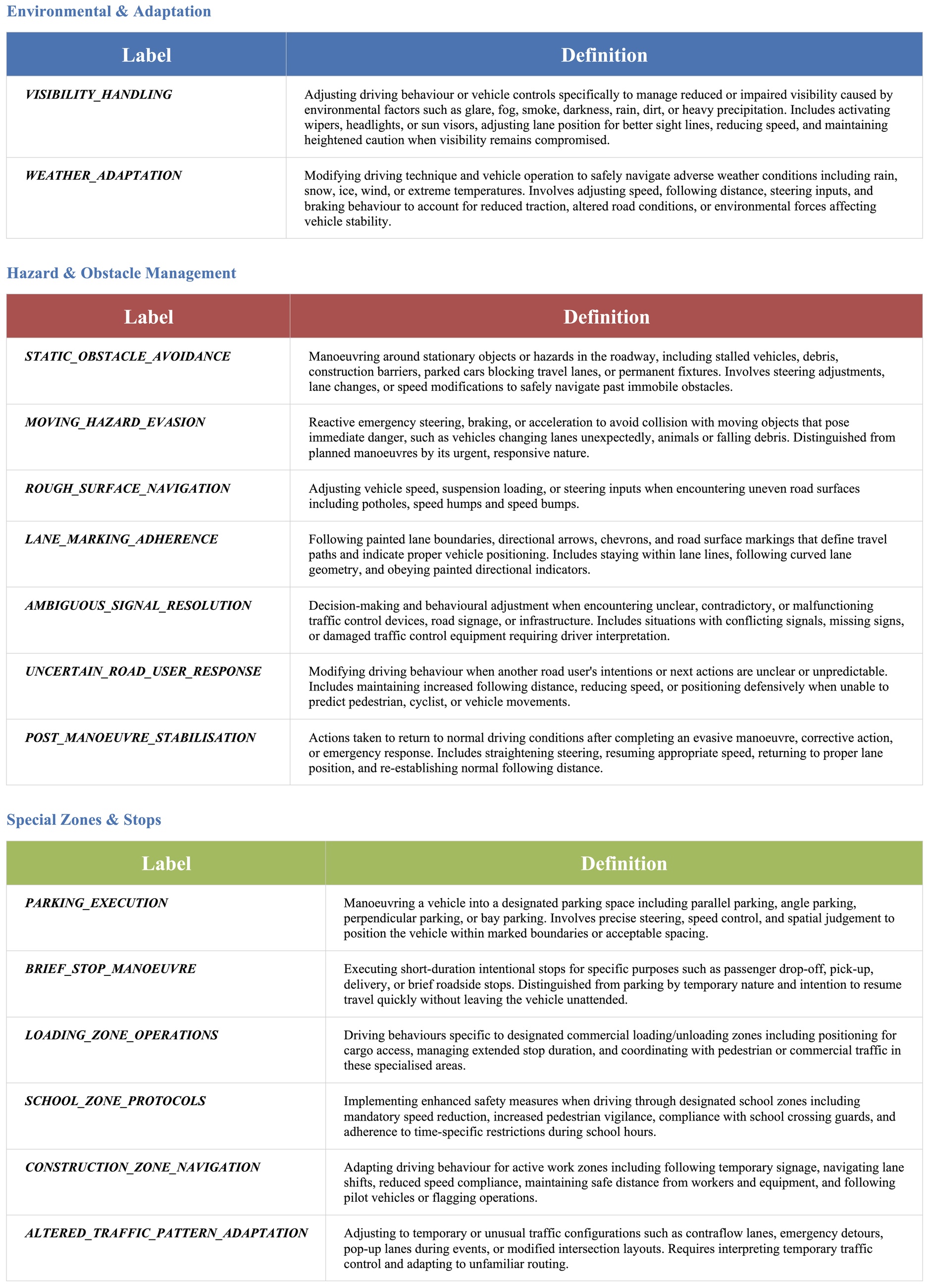}
\caption{Driving Context Taxonomy (Part 1).}
\label{fig:taxonomy_part1}
\end{figure}

\begin{figure}[!htbp]
\centering
\includegraphics[width=1.2\textwidth,height=0.8\textheight,keepaspectratio]{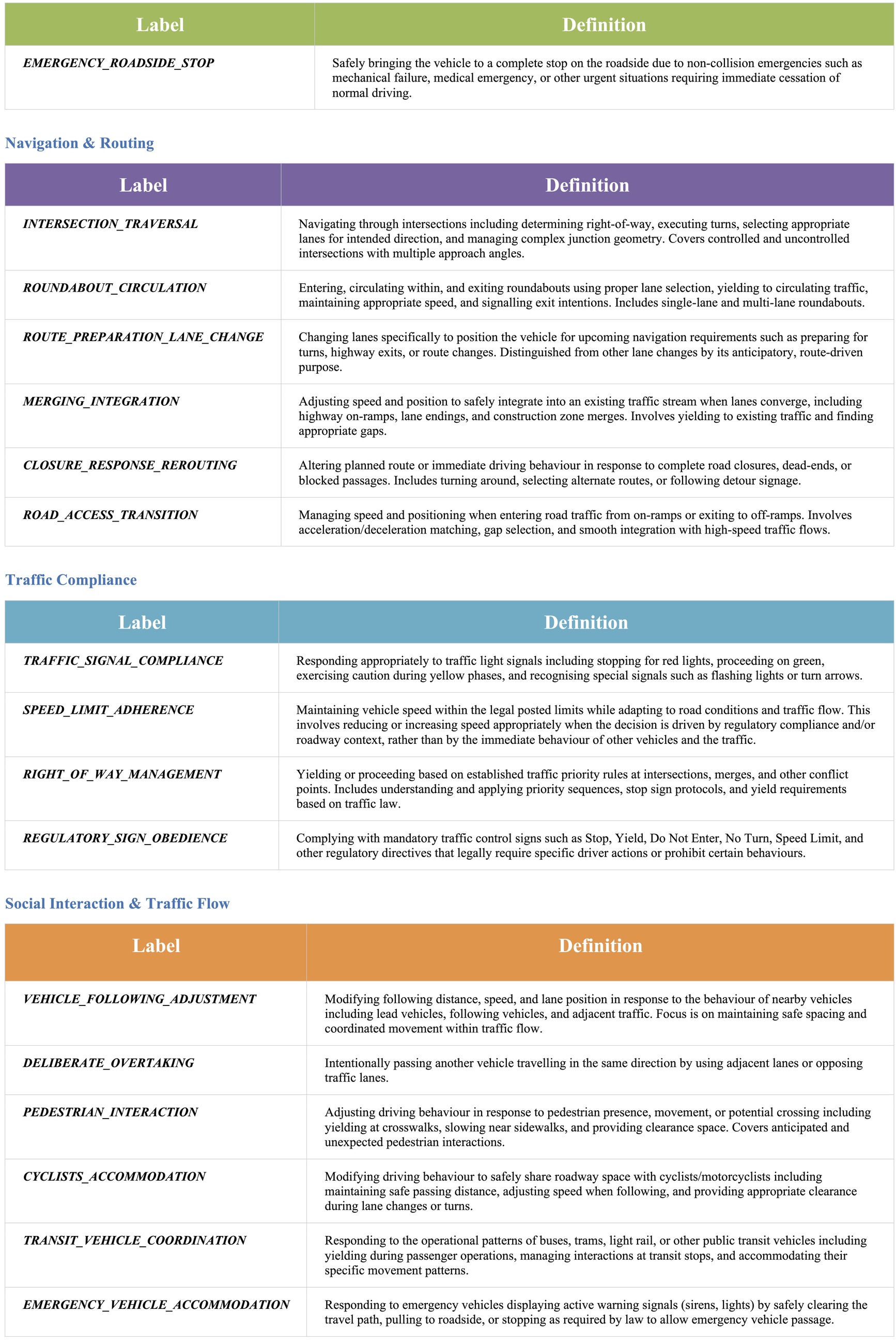}
\caption{Driving Context Taxonomy (Part 2).}
\label{fig:taxonomy_part2}
\end{figure}

\clearpage
\section{Representative Examples of Agreed Cases with Model Justifications}
\label{app:agreed_cases}

\begin{figure}[!htbp]
\centering
\includegraphics[width=1.1\textwidth,height=0.8\textheight,keepaspectratio]{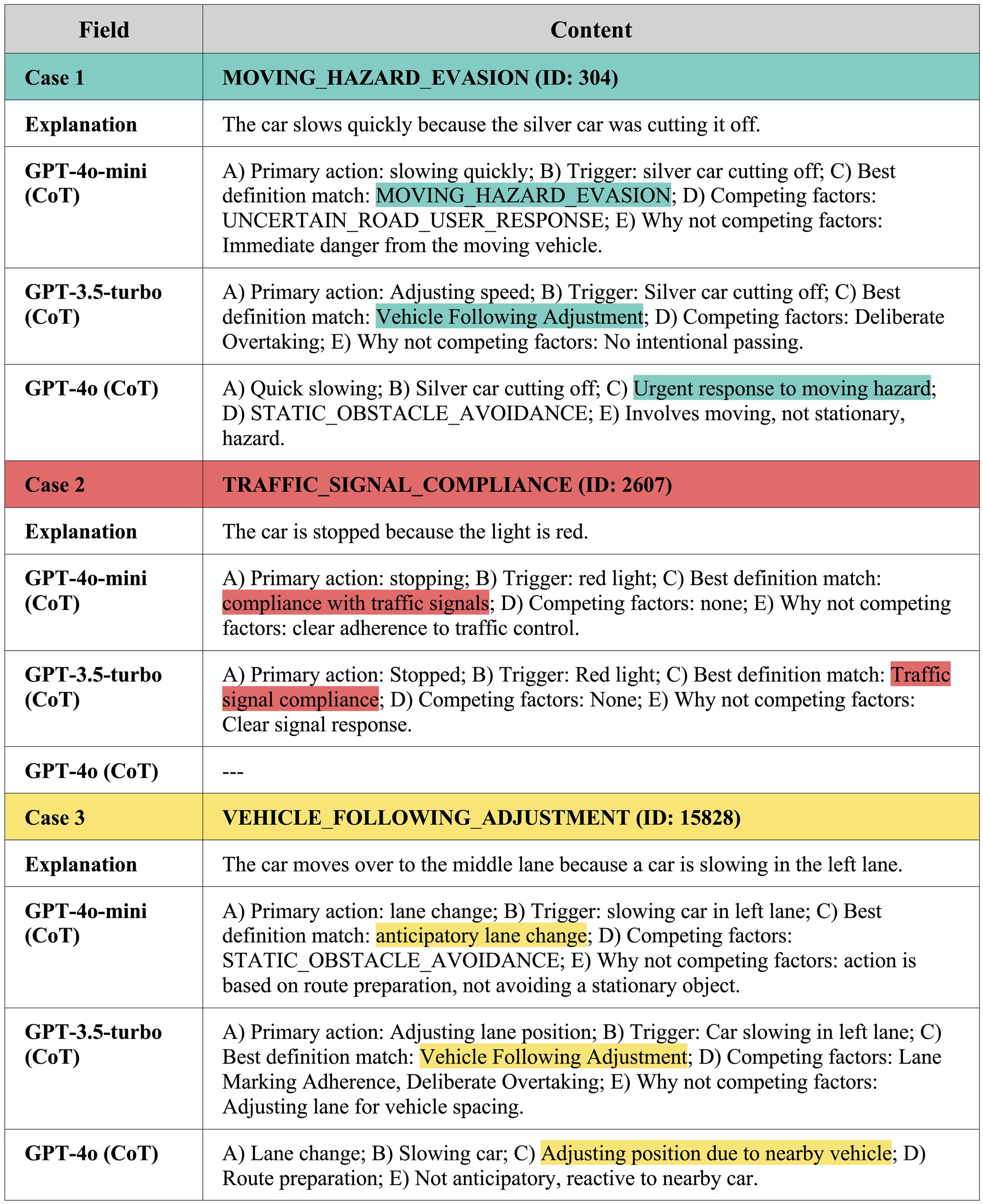}
\caption{Examples of model reasoning and justification paths for classification.}
\label{fig:appendix_table_A}
\end{figure}

\clearpage
\section{Prompt Structure Used for Classification}
\label{app:prompt}

\begin{figure}[!htbp]
\centering
\includegraphics[width=1.1\textwidth,height=0.8\textheight,keepaspectratio]{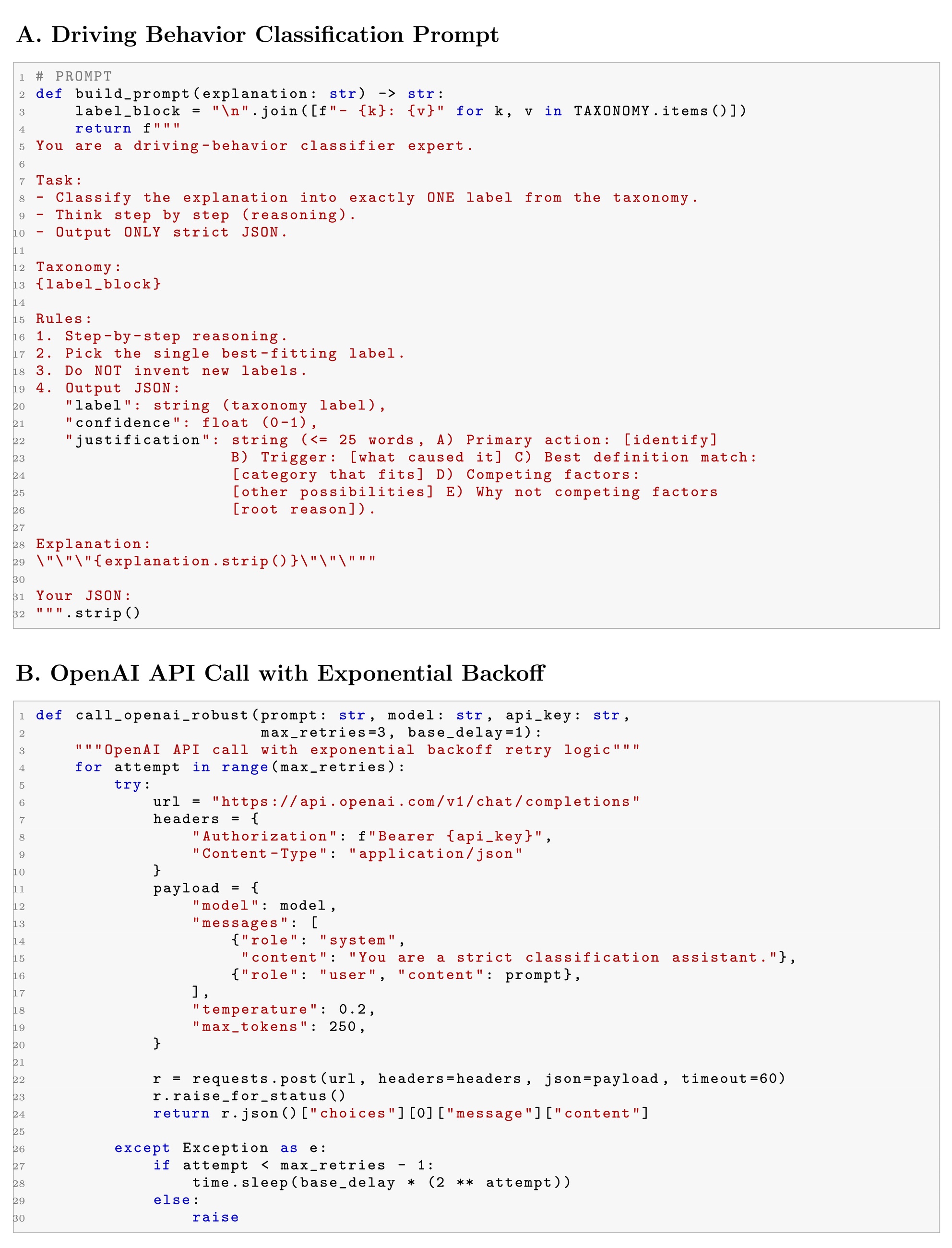}
\caption{Prompt structure used for driving context classification with the RACE model.}
\label{fig:prompts}
\end{figure}

\clearpage

\bibliographystyle{elsarticle-num}
\bibliography{references}

@article{cohen1960coefficient,
  title={A coefficient of agreement for nominal scales},
  author={Cohen, Jacob},
  journal={Educational and psychological measurement},
  volume={20},
  number={1},
  pages={37--46},
  year={1960},
  publisher={Sage Publications Sage CA: Thousand Oaks, CA}
}

@book{brezina2018statistics,
  title={Statistics in corpus linguistics: A practical guide},
  author={Brezina, Vaclav},
  year={2018},
  publisher={Cambridge University Press}
}

@incollection{artstein2017inter,
  title={Inter-annotator agreement},
  author={Artstein, Ron},
  booktitle={Handbook of linguistic annotation},
  pages={297--313},
  year={2017},
  publisher={Springer}
}

@article{brown2020language,
  title={Language models are few-shot learners},
  author={Brown, Tom and Mann, Benjamin and Ryder, Nick and Subbiah, Melanie and Kaplan, Jared D and Dhariwal, Prafulla and Neelakantan, Arvind and Shyam, Pranav and Sastry, Girish and Askell, Amanda and others},
  journal={Advances in neural information processing systems},
  volume={33},
  pages={1877--1901},
  year={2020}
}

@article{jiang2023llm,
  title={Llm-blender: Ensembling large language models with pairwise ranking and generative fusion},
  author={Jiang, Dongfu and Ren, Xiang and Lin, Bill Yuchen},
  journal={arXiv preprint arXiv:2306.02561},
  year={2023}
}

@article{faghihi2023role,
  title={The role of semantic parsing in understanding procedural text},
  author={Faghihi, Hossein Rajaby and Kordjamshidi, Parisa and Teng, Choh Man and Allen, James},
  journal={arXiv preprint arXiv:2302.06829},
  year={2023}
}

@article{chen2022traffic,
  title={Traffic accident duration prediction using text mining and ensemble learning on expressways},
  author={Chen, Jiaona and Tao, Weijun},
  journal={Scientific reports},
  volume={12},
  number={1},
  pages={21478},
  year={2022},
  publisher={Nature Publishing Group UK London}
}

@article{chowdhury2023applications,
  title={Applications of text mining in the transportation infrastructure sector: a review},
  author={Chowdhury, Sudipta and Alzarrad, Ammar},
  journal={Information},
  volume={14},
  number={4},
  pages={201},
  year={2023},
  publisher={MDPI}
}

@article{barmann2024incremental,
  title={Incremental learning of humanoid robot behavior from natural interaction and large language models},
  author={B{\"a}rmann, Leonard and Kartmann, Rainer and Peller-Konrad, Fabian and Niehues, Jan and Waibel, Alex and Asfour, Tamim},
  journal={Frontiers in Robotics and AI},
  volume={11},
  pages={1455375},
  year={2024},
  publisher={Frontiers Media SA}
}

@inproceedings{kollar2014grounding,
  title={Grounding verbs of motion in natural language commands to robots},
  author={Kollar, Thomas and Tellex, Stefanie and Roy, Deb and Roy, Nicholas},
  booktitle={Experimental robotics: The 12th international symposium on experimental robotics},
  pages={31--47},
  year={2014},
  organization={Springer}
}

@article{liu2024review,
  title={A review of natural-language-instructed robot execution systems},
  author={Liu, Rui and Guo, Yibei and Jin, Runxiang and Zhang, Xiaoli},
  journal={AI},
  volume={5},
  number={3},
  pages={948--989},
  year={2024},
  publisher={MDPI}
}

@article{graesser2016conversations,
  title={Conversations with AutoTutor help students learn},
  author={Graesser, Arthur C},
  journal={International Journal of Artificial Intelligence in Education},
  volume={26},
  number={1},
  pages={124--132},
  year={2016},
  publisher={Springer}
}

@article{mcenery2024corpus,
  title={Corpus linguistics and the social sciences},
  author={McEnery, Tony and Brookes, Gavin},
  journal={Corpus linguistics and linguistic theory},
  volume={20},
  number={3},
  pages={591--613},
  year={2024},
  publisher={De Gruyter}
}

@book{brookes2023corpus,
  title={Corpus linguistics for health communication: A guide for research},
  author={Brookes, Gavin and Collins, Luke C},
  year={2023},
  publisher={Routledge}
}

@book{mcenery2011corpus,
  title={Corpus linguistics: Method, theory and practice},
  author={McEnery, Tony and Hardie, Andrew},
  year={2011},
  publisher={Cambridge University Press}
}

@article{wei2022chain,
  title={Chain-of-thought prompting elicits reasoning in large language models},
  author={Wei, Jason and Wang, Xuezhi and Schuurmans, Dale and Bosma, Maarten and Xia, Fei and Chi, Ed and Le, Quoc V and Zhou, Denny and others},
  journal={Advances in neural information processing systems},
  volume={35},
  pages={24824--24837},
  year={2022}
}

@article{wang2022self,
  title={Self-consistency improves chain of thought reasoning in language models},
  author={Wang, Xuezhi and Wei, Jason and Schuurmans, Dale and Le, Quoc and Chi, Ed and Narang, Sharan and Chowdhery, Aakanksha and Zhou, Denny},
  journal={arXiv preprint arXiv:2203.11171},
  year={2022}
}

@article{lu2021relationship,
  title={The relationship between syntactic complexity and rhetorical move-steps in research article introductions: Variation among four social science and engineering disciplines},
  author={Lu, Xiaofei and Casal, J Elliott and Liu, Yingying and Kisselev, Olesya and Yoon, Jungwan},
  journal={Journal of English for Academic Purposes},
  volume={52},
  pages={101006},
  year={2021},
  publisher={Elsevier}
}

@article{lei2023large,
  title={A large-scale longitudinal study of syntactic complexity development in EFL writing: A mixed-effects model approach},
  author={Lei, Lei and Wen, Ju and Yang, Xiaohu},
  journal={Journal of Second Language Writing},
  volume={59},
  pages={100962},
  year={2023},
  publisher={Elsevier}
}

@article{gatt2018survey,
  title={Survey of the state of the art in natural language generation: Core tasks, applications and evaluation},
  author={Gatt, Albert and Krahmer, Emiel},
  journal={Journal of Artificial Intelligence Research},
  volume={61},
  pages={65--170},
  year={2018}
}

@book{biber2019register,
  title={Register, genre, and style},
  author={Biber, Douglas and Conrad, Susan},
  year={2019},
  publisher={Cambridge University Press}
}

@article{gries2021new,
  title={A new approach to (key) keywords analysis: Using frequency, and now also dispersion},
  author={Gries, Stefan Th},
  journal={Research in Corpus Linguistics},
  volume={9},
  number={2},
  pages={1--33},
  year={2021}
}

@article{singhal2023large,
  title={Large language models encode clinical knowledge},
  author={Singhal, Karan and Azizi, Shekoofeh and Tu, Tao and Mahdavi, S Sara and Wei, Jason and Chung, Hyung Won and Scales, Nathan and Tanwani, Ajay and Cole-Lewis, Heather and Pfohl, Stephen and others},
  journal={Nature},
  volume={620},
  number={7972},
  pages={172--180},
  year={2023},
  publisher={Nature Publishing Group}
}

@article{ding2023temporal,
  title={Temporal action segmentation: An analysis of modern techniques},
  author={Ding, Guodong and Sener, Fadime and Yao, Angela},
  journal={IEEE Transactions on Pattern Analysis and Machine Intelligence},
  volume={46},
  number={2},
  pages={1011--1030},
  year={2023},
  publisher={IEEE}
}

@article{achiam2023gpt,
  title={Gpt-4 technical report},
  author={Achiam, Josh and Adler, Steven and Agarwal, Sandhini and Ahmad, Lama and Akkaya, Ilge and Aleman, Florencia Leoni and Almeida, Diogo and Altenschmidt, Janko and Altman, Sam and Anadkat, Shyamal and others},
  journal={arXiv preprint arXiv:2303.08774},
  year={2023}
}

@inproceedings{nivre2016universal,
  title={Universal dependencies v1: A multilingual treebank collection},
  author={Nivre, Joakim and De Marneffe, Marie-Catherine and Ginter, Filip and Goldberg, Yoav and Hajic, Jan and Manning, Christopher D and McDonald, Ryan and Petrov, Slav and Pyysalo, Sampo and Silveira, Natalia and others},
  booktitle={Proceedings of the Tenth International Conference on Language Resources and Evaluation (LREC'16)},
  pages={1659--1666},
  year={2016}
}

@inproceedings{girju2003automatic,
  title={Automatic detection of causal relations for question answering},
  author={Girju, Roxana},
  booktitle={Proceedings of the ACL 2003 workshop on Multilingual summarization and question answering},
  pages={76--83},
  year={2003}
}

@incollection{prasad2017penn,
  title={The Penn Discourse Treebank: An annotated corpus of discourse relations},
  author={Prasad, Rashmi and Webber, Bonnie and Joshi, Aravind},
  booktitle={Handbook of linguistic annotation},
  pages={1197--1217},
  year={2017},
  publisher={Springer}
}

@inproceedings{de2008stanford,
  title={The Stanford typed dependencies representation},
  author={De Marneffe, Marie-Catherine and Manning, Christopher D},
  booktitle={Coling 2008: proceedings of the workshop on cross-framework and cross-domain parser evaluation},
  pages={1--8},
  year={2008}
}

@article{monroe2008fightin,
  title={Fightin'words: Lexical feature selection and evaluation for identifying the content of political conflict},
  author={Monroe, Burt L and Colaresi, Michael P and Quinn, Kevin M},
  journal={Political Analysis},
  volume={16},
  number={4},
  pages={372--403},
  year={2008},
  publisher={Cambridge University Press}
}

@book{gries2016quantitative,
  title={Quantitative corpus linguistics with R: A practical introduction},
  author={Gries, Stefan Th},
  year={2016},
  publisher={Routledge}
}

@inproceedings{eisenstein2011discovering,
  title={Discovering sociolinguistic associations with structured sparsity},
  author={Eisenstein, Jacob and Smith, Noah A and Xing, Eric},
  booktitle={Proceedings of the 49th annual meeting of the association for computational linguistics: human language technologies},
  pages={1365--1374},
  year={2011}
}

@article{honnibal2020spacy,
  title={spaCy: Industrial-strength natural language processing in python},
  author={Honnibal, Matthew and Montani, Ines and Van Landeghem, Sofie and Boyd, Adriane and others},
  year={2020},
  publisher={Zenodo, Honolulu, HI, USA}
}

@book{vajjala2020practical,
  title={Practical natural language processing: a comprehensive guide to building real-world NLP systems},
  author={Vajjala, Sowmya and Majumder, Bodhisattwa and Gupta, Anuj and Surana, Harshit},
  year={2020},
  publisher={O'Reilly Media}
}

@article{fleiss1971measuring,
  title={Measuring nominal scale agreement among many raters.},
  author={Fleiss, Joseph L},
  journal={Psychological bulletin},
  volume={76},
  number={5},
  pages={378},
  year={1971},
  publisher={American Psychological Association}
}

@article{landis1977measurement,
  title={The measurement of observer agreement for categorical data},
  author={Landis, J Richard and Koch, Gary G},
  journal={biometrics},
  pages={159--174},
  year={1977},
  publisher={JSTOR}
}

@book{alammar2024hands,
  title={Hands-on large language models: language understanding and generation},
  author={Alammar, J. and Grootendorst, M.},
  year={2024},
  publisher={O'Reilly Media, Inc.}
}

@article{albrecht2022despite,
  title={Despite "super-human" performance, current LLMs are unsuited for decisions about ethics and safety},
  author={Albrecht, Joshua and Kitanidis, Ellie and Fetterman, Abraham J.},
  journal={arXiv preprint arXiv:2212.06295},
  year={2022}
}

@article{baby2024development,
  title={Development and classification of autonomous vehicle's ambiguous driving scenario},
  author={Baby, Teena and Ippoliti, Hande {\c{S}}. and Wintersberger, Philipp and Zhang, Yichen and Yoon, Sang Ho and Lee, Joonwoo and Lee, Seul Chan},
  journal={Accident Analysis \& Prevention},
  volume={200},
  pages={107501},
  year={2024}
}

@article{vaswani2017attention,
  title={Attention is all you need},
  author={Vaswani, Ashish and Shazeer, Noam and Parmar, Niki and Uszkoreit, Jakob and Jones, Llion and Gomez, Aidan N and Kaiser, {\L}ukasz and Polosukhin, Illia},
  journal={Advances in neural information processing systems},
  volume={30},
  year={2017}
}

@article{zhao2023survey,
  title={A survey of large language models},
  author={Zhao, Wayne Xin and Zhou, Kun and Li, Junyi and Tang, Tianyi and Wang, Xiaolei and Hou, Yupeng and Min, Yingqian and Zhang, Beichen and Zhang, Junjie and Dong, Zican and others},
  journal={arXiv preprint arXiv:2303.18223},
  volume={1},
  number={2},
  year={2023}
}

@article{minaee2024large,
  title={Large language models: A survey},
  author={Minaee, Shervin and Mikolov, Tomas and Nikzad, Narjes and Chenaghlu, Meysam and Socher, Richard and Amatriain, Xavier and Gao, Jianfeng},
  journal={arXiv preprint arXiv:2402.06196},
  year={2024}
}

@article{chen2023driving,
  title={Driving with LLMs: Fusing object-level vector modality for explainable autonomous driving},
  author={Chen, Long and Sinavski, Oleg and H{\"u}nermann, Jan and Karnsund, Alice and Willmott, Andrew James and Birch, Danny and Maund, Daniel and Shotton, Jamie},
  journal={arXiv preprint arXiv:2310.01957},
  year={2023}
}

@article{chen2024endtoend,
  title={End-to-end autonomous driving: Challenges and frontiers},
  author={Chen, Li and Wu, Penghao and Chitta, Kashyap and Jaeger, Bernhard and Geiger, Andreas and Li, Hongyang},
  journal={IEEE Transactions on Pattern Analysis and Machine Intelligence},
  year={2024}
}

@article{chen2023feature,
  title={Feature selection for driving style and skill clustering using naturalistic driving data and driving behavior questionnaire},
  author={Chen, Yonggang and Wang, Kui and Lu, Jian John},
  journal={Accident Analysis \& Prevention},
  volume={185},
  pages={107022},
  year={2023}
}

@article{cui2024receive,
  title={Receive, reason, and react: Drive as you say, with large language models in autonomous vehicles},
  author={Cui, Can and Ma, Yunsheng and Cao, Xu and Ye, Wenqian and Wang, Ziran},
  journal={IEEE Intelligent Transportation Systems Magazine},
  volume={16},
  number={4},
  pages={81--94},
  year={2024}
}

@techreport{degelder2020scenario,
  title={Scenario categories for the assessment of automated vehicles},
  author={de Gelder, Erwin and den Camp, Olaf Op and de Boer, Niek},
  institution={CETRAN, Singapore},
  year={2020},
  type={Version 1}
}

@article{mehraban2025saliency,
  title={Saliency-Guided Domain Adaptation for Left-Hand Driving in Autonomous Steering},
  author={Mehraban, Zahra and Glaser, Sebastien and Milford, Michael and Schroeter, Ronald},
  journal={arXiv preprint arXiv:2511.01223},
  year={2025}
}

@article{ding2023survey,
  title={A survey on safety-critical driving scenario generation—a methodological perspective},
  author={Ding, Wenhao and Xu, Chejian and Arief, Mansur and Lin, Haohong and Li, Bo and Zhao, Ding},
  journal={IEEE Transactions on Intelligent Transportation Systems},
  volume={24},
  number={7},
  pages={6971--6988},
  year={2023}
}

@article{dong2023why,
  title={Why did the AI make that decision? Towards an explainable artificial intelligence (XAI) for autonomous driving systems},
  author={Dong, Jingxuan and Chen, Sikai and Miralinaghi, Mohammad and Chen, Tianle and Li, Peibo and Labi, Samuel},
  journal={Transportation Research Part C: Emerging Technologies},
  volume={156},
  pages={104358},
  year={2023}
}

@inproceedings{feng2023driving,
  title={Driving style classification using deep temporal clustering with enhanced explainability},
  author={Feng, Yimin and Ye, Qiyang and Adan, Furkan and Marques, Lino and Angeloudis, Panagiotis},
  booktitle={2023 IEEE 26th International Conference on Intelligent Transportation Systems (ITSC)},
  year={2023}
}

@article{giamattei2024causality,
  title={Causality-driven testing of autonomous driving systems},
  author={Giamattei, Luca and Guerriero, Antonio and Pietrantuono, Roberto and Russo, Stefano},
  journal={ACM Transactions on Software Engineering and Methodology},
  volume={33},
  number={3},
  pages={1--35},
  year={2024}
}

@inproceedings{huang2024driverp,
  title={DriverP: RAG and prompt engineering embodied parallel driving in cyber-physical-social spaces},
  author={Huang, Jingxuan and Ma, Huaqing and Zhang, Tianxiang and Lin, Feng and Ma, Shuo and Wang, Xiao and Wang, Fei-Yue},
  booktitle={2024 IEEE 4th International Conference on Digital Twins and Parallel Intelligence (DTPI)},
  year={2024}
}

@article{jin2023adapt,
  title={ADAPT: Action-aware Driving Caption Transformer},
  author={Jin, Bu and Liu, Xinyu and Zheng, Yupeng and Li, Pengfei and Zhao, Hao and Zhang, Tong and Zheng, Yuhang and Zhou, Guyue and Liu, Jingjing},
  year={2023}
}

@inproceedings{kim2018textual,
  title={Textual explanations for self-driving vehicles},
  author={Kim, Jinkyu and Rohrbach, Anna and Darrell, Trevor and Canny, John and Akata, Zeynep},
  booktitle={Proceedings of the European Conference on Computer Vision (ECCV)},
  year={2018}
}

@article{kuznietsov2024explainable,
  title={Explainable AI for safe and trustworthy autonomous driving: A systematic review},
  author={Kuznietsov, Artem and Gyevnar, Balint and Wang, Cheng and Peters, Steven and Albrecht, Stefano V.},
  journal={arXiv preprint arXiv:2402.10086},
  year={2024}
}

@article{lee2024applying,
  title={Applying large language models and chain-of-thought for automatic scoring},
  author={Lee, Gyeong-Geon and Latif, Ehsan and Wu, Xuanlong and Liu, Ning and Zhai, Xiaoming},
  journal={Computers and Education: Artificial Intelligence},
  volume={6},
  pages={100213},
  year={2024}
}

@article{li2024exploring,
  title={Exploring the causality of end-to-end autonomous driving},
  author={Li, Jiankun and Li, Hao and Liu, Jiaming and Zou, Zhifeng and Ye, Xiaolong and Wang, Fanyi and Huang, Junbo and Wu, Haiming and Wang, Hongxiang},
  journal={arXiv preprint arXiv:2407.06546},
  year={2024}
}

@article{liao2023ai,
  title={AI transparency in the age of LLMs: A human-centered research roadmap},
  author={Liao, Q. Vera and Vaughan, Jennifer Wortman},
  journal={arXiv preprint arXiv:2306.01941},
  pages={5368--5393},
  year={2023}
}

@inproceedings{luo2024pkrdcot,
  title={PKRD-CoT: A unified chain-of-thought prompting for multi-modal large language models in autonomous driving},
  author={Luo, Xiang and Ding, Feiyang and Song, Yang and Zhang, Xiaoqing and Loo, Jonathan},
  booktitle={International Conference on Neural Information Processing},
  year={2024}
}

@article{mao2023gptdriver,
  title={GPT-driver: Learning to drive with GPT},
  author={Mao, Jiageng and Qian, Yuxi and Zhao, Hang and Wang, Yue},
  journal={arXiv preprint arXiv:2310.01415},
  year={2023}
}

@article{mao2023language,
  title={A language agent for autonomous driving},
  author={Mao, Jiageng and Ye, Junjie and Qian, Yuxi and Pavone, Marco and Wang, Yue},
  journal={arXiv preprint arXiv:2311.10813},
  year={2023}
}

@article{mehraban2024fuzzy,
  title={Fuzzy adaptive cruise control with model predictive control responding to dynamic traffic conditions for automated driving},
  author={Mehraban, Zahra and Zadeh, Abbas Yousefi and Khayyam, Hamid and Mallipeddi, Rammohan and Jamali, Ali},
  journal={Engineering Applications of Artificial Intelligence},
  volume={136},
  pages={109008},
  year={2024}
}

@article{nascimento2020systematic,
  title={A systematic literature review about the impact of artificial intelligence on autonomous vehicle safety},
  author={Nascimento, Adriano M. and Vismari, Lucio F. and Molina, Cristiane B. S. T. and Cugnasca, Paulo S. and Camargo, Jo{\~a}o B. and Almeida, Jorge R. De and Inam, Rafia and Fersman, Elena and Marquezini, Marcelo V. and Hata, Alberto Y.},
  journal={IEEE Transactions on Intelligent Transportation Systems},
  volume={21},
  number={12},
  pages={4928--4946},
  year={2020},
  doi={10.1109/TITS.2019.2949915}
}

@article{pan2025chatleafdisease,
  title={ChatLeafDisease: A chain-of-thought prompting approach for crop disease classification using large language models},
  author={Pan, Jianwu and Zhong, Ruoling and Xia, Faming and Huang, Jinyang and Zhu, Lin and Yang, Yang and Lin, Tao},
  journal={Plant Phenomics},
  pages={100094},
  year={2025}
}

@article{sha2023languagempc,
  title={LanguageMPC: Large language models as decision makers for autonomous driving},
  author={Sha, Hao and Mu, Yao and Jiang, Yuxuan and Chen, Lirui and Xu, Chenfeng and Luo, Ping and Li, Shengbo Eben and Tomizuka, Masayoshi and Zhan, Wei and Ding, Mingyu},
  journal={arXiv preprint arXiv:2310.03026},
  year={2023}
}

@techreport{taxonomy2017,
  title={Taxonomy of scenarios for automated driving},
  author={{Connected Places Catapult}},
  year={2017},
  url={https://cp-catapult.s3.amazonaws.com/uploads/2021/07/ATS34-Taxonomy-of-Scenarios-for-Automated-Driving-Report.pdf}
}

@article{tekkesinoglu2024advancing,
  title={Advancing explainable autonomous vehicle systems: A comprehensive review and research roadmap},
  author={Tekkesinoglu, Sena and Habibovic, Azra and Kunze, Lars},
  journal={arXiv preprint arXiv:2404.00019},
  year={2024}
}

@book{hilpert2019construction,
  title={Construction grammar and its application to English},
  author={Hilpert, Martin},
  year={2019},
  publisher={Edinburgh University Press}
}

@article{voneschenbach2021transparency,
  title={Transparency and the black box problem: Why we do not trust AI},
  author={Von Eschenbach, William J.},
  journal={Philosophy \& Technology},
  volume={34},
  number={4},
  pages={1607--1622},
  year={2021},
  doi={10.1007/s13347-021-00477-0}
}

@article{wang2024soft,
  title={Soft self-consistency improves language model agents},
  author={Wang, Han and Prasad, Archiki and Stengel-Eskin, Elias and Bansal, Mohit},
  journal={arXiv preprint arXiv:2402.13212},
  year={2024}
}

@article{wang2023goaldriven,
  title={Goal-driven explainable clustering via language descriptions},
  author={Wang, Zihan and Shang, Jingbo and Zhong, Ruiqi},
  journal={arXiv preprint arXiv:2305.13749},
  year={2023}
}

@article{wen2023ontheroad,
  title={On the road with GPT-4V(ision): Early explorations of visual-language model on autonomous driving},
  author={Wen, Licheng and Yang, Xuemeng and Fu, Daocheng and Wang, Xiaofeng and Cai, Pinlong and Li, Xin and Ma, Tao and Li, Yingxuan and Xu, Linran and Shang, Dengke},
  journal={arXiv preprint arXiv:2311.05332},
  year={2023}
}

@article{xu2023drivegpt4,
  title={DriveGPT4: Interpretable end-to-end autonomous driving via large language model},
  author={Xu, Zhenhua and Zhang, Yujia and Xie, Enze and Zhao, Zijian and Guo, Yong and Wong, Kenneth K. and Li, Zhenguo and Zhao, Hengshuang},
  journal={arXiv preprint arXiv:2310.01412},
  year={2023}
}

@article{yang2023llm4drive,
  title={LLM4Drive: A survey of large language models for autonomous driving},
  author={Yang, Zhenjie and Jia, Xiaosong and Li, Hongyang and Yan, Junchi},
  journal={arXiv preprint arXiv:2311.01043},
  year={2023}
}

@inproceedings{yousefizadeh2025psylingxav,
  title={PsyLingXAV: A psycholinguistics design framework for XAI in automated vehicles},
  author={Yousefi Zadeh, Abbas and Li, Xiang and Rakotonirainy, Andry and Schroeter, Ronald and Glaser, S{\'e}bastien},
  booktitle={Joint Proceedings of the xAI 2025 Late-breaking Work, Demos and Doctoral Consortium: co-located with the 3rd World Conference on eXplainable Artificial Intelligence (xAI 2025): Istanbul, Turkey, July 09--11, 2025},
  year={2025}
}

@inproceedings{yu2020bdd100k,
  title={BDD100K: A diverse driving dataset for heterogeneous multitask learning},
  author={Yu, Fisher and Chen, Haofeng and Wang, Xin and Xian, Wenqi and Chen, Yingying and Liu, Fangchen and Madhavan, Vashisht and Darrell, Trevor},
  booktitle={Proceedings of the IEEE/CVF Conference on Computer Vision and Pattern Recognition},
  year={2020}
}

@article{yuan2024ragdriver,
  title={RAG-Driver: Generalisable driving explanations with retrieval-augmented in-context learning in multi-modal large language model},
  author={Yuan, Jianhao and Sun, Shuyang and Omeiza, Daniel and Zhao, Bo and Newman, Paul and Kunze, Lars and Gadd, Matthew},
  journal={arXiv preprint arXiv:2402.10828},
  year={2024}
}

@article{zadeh2024integrated,
  title={Integrated intelligent control systems for eco and safe driving in autonomous vehicles},
  author={Zadeh, Abbas Yousefi and Khayyam, Hamid and Mallipeddi, Rammohan and Jamali, Ali},
  journal={IEEE Transactions on Intelligent Transportation Systems},
  year={2024}
}

@article{zhao2024risk,
  title={Risk scenario generation for autonomous driving systems based on causal Bayesian networks},
  author={Zhao, Jinghan and Du, Delong and Yu, Xiaotong and Li, Hongming},
  journal={arXiv preprint arXiv:2405.16063},
  year={2024}
}

@article{zhen2024leveraging,
  title={Leveraging large language models with chain-of-thought and prompt engineering for traffic crash severity analysis and inference},
  author={Zhen, Hao and Shi, Yanbing and Huang, Yang and Yang, John Junye and Liu, Ning},
  journal={Computers},
  volume={13},
  number={9},
  pages={232},
  year={2024}
}

@article{zhu2025humancentric,
  title={Human-centric explanations for users in automated vehicles: A systematic review},
  author={Zhu, Zhuoren and Li, Xiang and Delhomme, Patricia and Schroeter, Ronald and Glaser, S{\'e}bastien and Rakotonirainy, Andry},
  journal={Accident Analysis \& Prevention},
  volume={220},
  pages={108152},
  year={2025}
}

\end{document}